\documentclass[11pt]{article}

\usepackage[preprint]{acl}

\usepackage{times}
\usepackage{latexsym}
\usepackage{multirow}
\usepackage{stmaryrd}
\usepackage[T1]{fontenc}

\usepackage[utf8]{inputenc}

\usepackage{microtype}

\usepackage{graphicx}
\usepackage{subcaption}
\usepackage{inconsolata}

\usepackage{graphicx}

\usepackage{amsmath,amssymb,amsthm}
\usepackage{booktabs}
\usepackage{microtype}
\usepackage{parskip}
\usepackage{natbib}
\usepackage{hyperref}
\usepackage{xcolor}
\usepackage{enumitem}
\usepackage{titlesec}
\usepackage{graphicx}
\usepackage{amsmath}
\usepackage{algorithm}
\usepackage{algpseudocode}

\usepackage{tikz}
\usetikzlibrary{arrows.meta,positioning,fit,calc,shapes.multipart}

\newcommand{\op}[1]{\textsc{#1}}

\title{Is Inference Mediated by Distinct Semantic Structures in LLMs?\\A Mechanistic Interpretation}
\author{Nura Aljaafari$^{1}$, Marco Valentino$^{2}$, Andr\'{e} Freitas$^{1,3,4}$ \\
  $^{1}$ University of Manchester, United Kingdom, 
  $^{2}$ University of Sheffield, United Kingdom\\
  $^{3}$ Idiap Research Institute, Switzerland\\
  $^{4}$ CRUK National Biomarker Centre, University of Manchester, United Kingdom\\
  \texttt{\{nura.aljaafari,andre.freitas\}@manchester.ac.uk} \\
  \texttt{m.valentino@sheffield.ac.uk}}
\date{}

\begin{document}
\maketitle

\begin{abstract}
Predicting a label correctly does not necessarily require representing the operation that produces it. Transformer representations are known to carry label-level information, but whether they encode semantic operations producing those labels is unclear. We investigate this in Natural Language Inference using controlled premise-hypothesis pairs that differ by a single semantic transformation. Using layer-wise activations, we estimate operation-level subspaces via SVD and test their causal relevance through activation steering in four open-weight decoder models. Transformation effects are decodable with $84.8$-$99\%$ accuracy and occupy partially distinct but overlapping subspaces, exceeding random-subspace baselines. Steering experiments show that these directions causally influence predictions, though steerability varies across models; cross-operation steering further reveals structured interference and a dissociation between subspace selectivity and cross-operation independence. These findings indicate that the models encode not only \emph{that} a hypothesis relates to a premise but also, in part, \emph{how} it does so, implying that mechanistic analysis and control should operate at the level of semantic operations rather than predicted labels alone.\footnote{Code and supplementary materials are available at \url{[anonymised for review]}.}
\end{abstract}
\section{Introduction}\label{sec:introduction}
Understanding how transformer models represent semantic relationships requires going beyond what a model predicts to how its internal computation is organised \citep{aljaafari-etal-2026-llms,olah2020zoom}. Since natural language inference (NLI) labels can result from different underlying semantic operations \citep{bowman2015snli,williams2018multinli}, it provides a controlled setting for asking whether models represent only the final relation label or also the transformation that produces it (see App.~\ref{sec:appendix_foundations} for a formal linguistic and logical grounding of this distinction); a question with direct consequences for mechanistic interpretability, evaluation, and control \citep{geiger2021causal}.

\begin{figure}
    \centering
    \includegraphics[width=\linewidth]{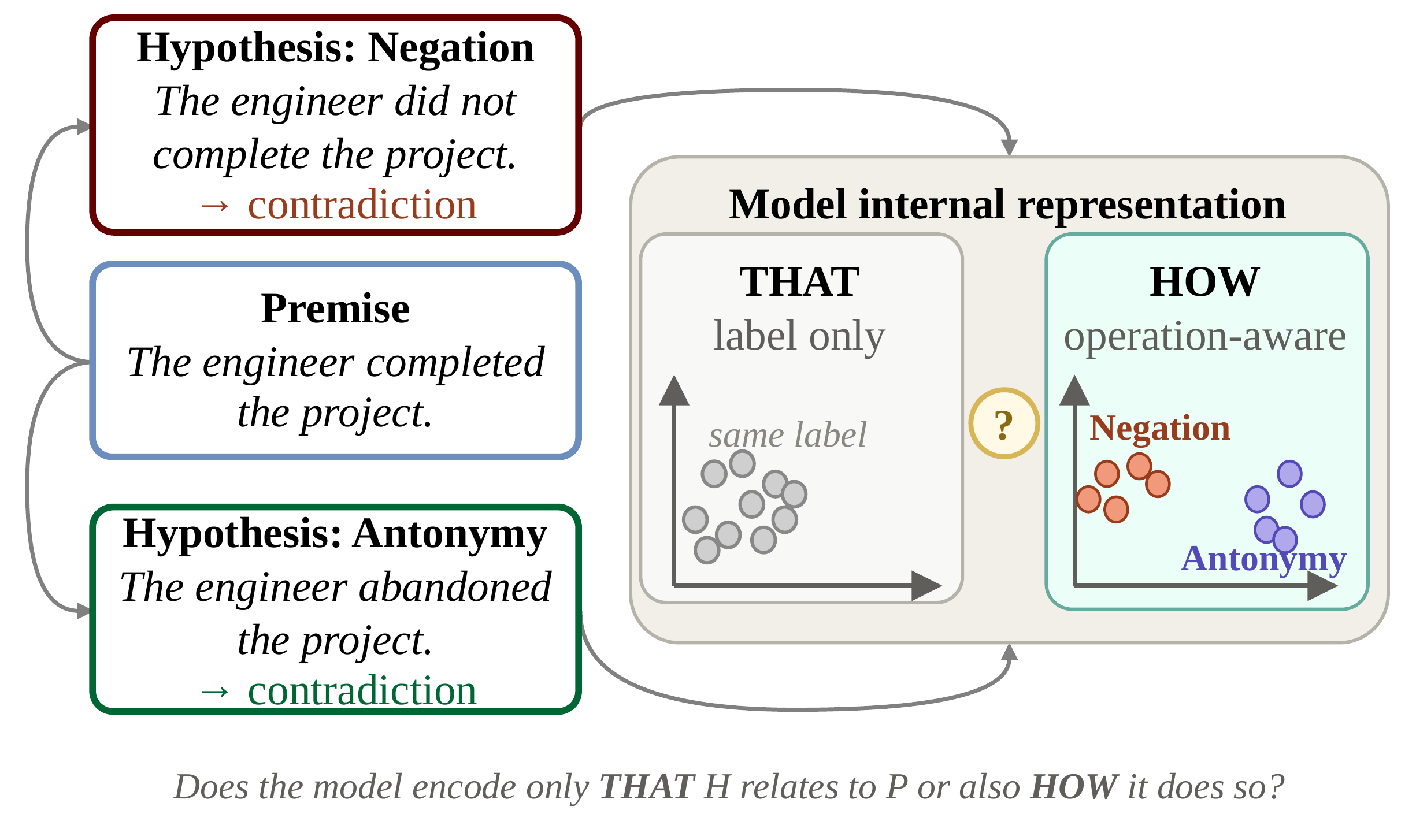}
    \caption{\textbf{Transformation-level ambiguity in entailment representations.} A single premise can be paired with hypotheses produced by different semantic operations. We ask whether the model represents only the shared label or also the transformation that produces it.}
    \label{fig:nli_example}
\end{figure}

\begin{figure*}[ht]
\centering
    \includegraphics[width=.9\textwidth]{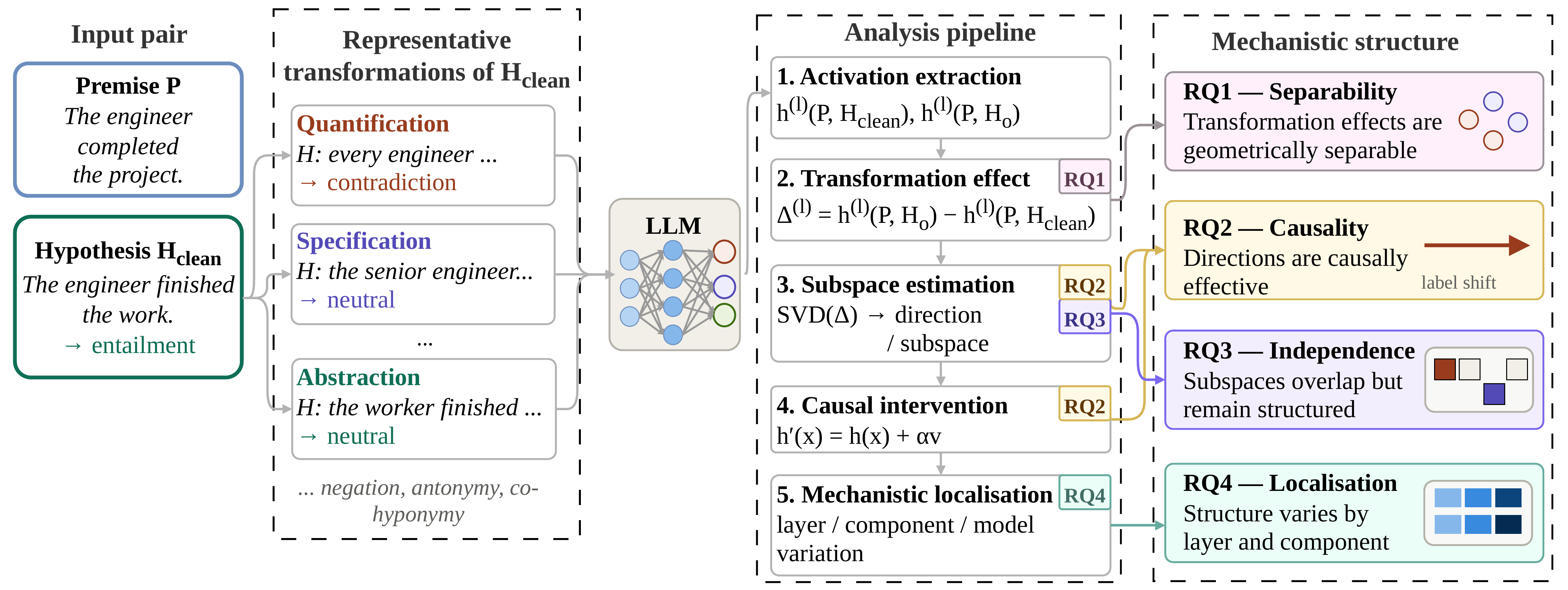}
    \caption{\textbf{Method overview.} Controlled transformations are applied to the hypothesis in a premise-hypothesis pair. We extract activation differences $\Delta^{(l)} = h^{(l)}(P, H_o) - h^{(l)}(P, H_{\text{clean}})$ at each layer, estimate operation-specific subspaces $\mathcal{S}_o^{(l)}$ using SVD, and test derived directions through bidirectional activation steering. RQ labels indicate which analysis each stage supports.}
    \label{fig:methodology_overview}
\end{figure*}

Fig.~\ref{fig:nli_example} illustrates this ambiguity: the premise \emph{The engineer completed the project}, both negation \emph{The engineer did not complete the project}, and antonymy \emph{The engineer abandoned the project} produce a contradiction label despite involving distinct semantic operations. Since the training objective does not directly require these operations to be distinguished, the shared label may obscure whether their internal representations are transformation-specific.

Existing work motivates this question from several directions, but does not address it directly. Probing studies show that semantic and relational properties are often linearly decodable from hidden states \citep{tenney2019bert,hewitt2019structural,belinkov2022probing}, but decodability alone does not establish that the model uses this information \citep{hewitt2019designing,voita2020information}. Interventional work on NLI has identified causally relevant natural-logic features \citep{geiger2020neural,rozanova-etal-2023-interventional,rozanova-etal-2024-estimating}, while representation-geometry and activation-steering studies show that semantic directions can be both geometrically consistent and causally influential \citep{park2024linear,hernandez2024linearity,marks2024geometry,turner2023activation,zou2023representation}. However, these lines of work do not establish whether a \emph{class} of meaning-altering operations shares low-dimensional geometric structure, or whether that structure causally contributes to NLI predictions.

To investigate this, we construct a controlled 1,800-pair contrastive NLI dataset in which each clean premise-hypothesis input is paired with a semantic variant differing by a single meaning-altering operation, spanning six operations in total (Fig.~\ref{fig:methodology_overview}). This pairing isolates transformation-induced activation differences, which we analyse through four questions: \textbf{(RQ1) \emph{Separability:}} whether activation differences from distinct transformation types occupy geometrically distinguishable regions; \textbf{(RQ2) \emph{Causality:}} whether directions derived from these differences systematically shift model predictions under targeted interventions; \textbf{(RQ3) \emph{Independence:}} how subspace selectivity relates to cross-operation interference, and whether these properties are dissociable; and \textbf{(RQ4) \emph{Localisation:}} at which layers and components transformation-dependent structure emerges and label-level evidence is amplified. We study six transformations, Negation, Antonymy, Quantification, Abstraction, Specification, and Co-hyponymy, across four open-weight decoder models at the 1.7-4B scale.

The results provide mechanistic evidence that models encode not only label outcomes but also, in part, the operations that produce them. \emph{Transformation effects are linearly decodable with $84.8$-$99\%$ accuracy across models}, well above random subspace baselines, and \emph{activation steering confirms their causal relevance in three of four models}. Operations differ in how selectively their effects are concentrated within their own subspaces, and \emph{this selectivity is dissociable from cross-operation independence}: an operation with moderate subspace selectivity can produce stronger cross-operation interference than the operation with the highest selectivity. These findings indicate that \emph{label-level analyses undercharacterise model behaviour}, and that \emph{the relevant structure is defined at the level of semantic operations rather than outputs alone}. We summarise our contributions as follows: 

\begin{itemize}[leftmargin=*,itemsep=1pt]
    \item A mechanistic framework combining contrastive transformation-pair construction, operation-specific subspace identification, bidirectional activation steering, and logit attribution to analyse semantic transformations as geometric effects in model representations.
    \item Empirical evidence that transformation activation differences are \emph{linearly decodable} with high accuracy and \emph{causally steerable}, showing that models encode not only label-level outcomes but also, in part, the operations that lead to them.
    \item Evidence of a dissociation between subspace selectivity and cross-operation independence, and an asymmetry between forward and inverse steering, both constraining mechanistic accounts of transformation-level representation.
\end{itemize}

\section{Related Work}
\label{sec:related_work}
\paragraph{Probing and representation analysis.} Probing studies show that syntactic, semantic, and relational properties are decodable from transformer representations \citep{tenney2019bert,jawahar2019does,hewitt2019structural}, but a successful probe shows that information is \emph{present}, not that the model \emph{uses} it; control-task and information-theoretic refinements sharpen the methodology without closing this gap \citep{belinkov2022probing,hewitt2019designing,voita2020information}. Our work adds causal evidence: transformation-level structure is not only decodable but behaviourally active under targeted steering.

\paragraph{Mechanistic interpretability.} The circuits framework \citep{olah2020zoom,elhage2021mathematical} decomposes transformer computation into interpretable subgraphs; subsequent work has identified components associated with indirect object identification \citep{wang2022interpretability}, modular arithmetic \citep{nanda2023progress}, and syntactic agreement \citep{finlayson2021causal}, with automated discovery \citep{conmy2023automated} and causal abstraction \citep{geiger2021causal,wu2023interpretability} scaling these analyses. Rather than identifying circuits that produce a label, we characterise whether \emph{transformation-level effects} form structured geometric objects in activation space that can be isolated and causally manipulated.

\paragraph{Activation steering and representation engineering.} Contrastive activation addition \citep{turner2023activation, rimsky2024steering}, inference-time intervention \citep{li2023inference}, representation engineering \citep{zou2023representation}, and contrast-consistent search \citep{burns2023discovering} show that semantic properties correspond to steerable directions in representation space; \citet{zhao_when_2025} further demonstrate that SVD-identified subspaces can be ablated to disentangle surface features from reasoning, a close methodological precedent of our approach. We extend this paradigm from broad behavioural properties to six fine-grained semantic operations, enabling direct comparison of transformation-specific directions and their asymmetric induction and reversal behaviour.

\paragraph{NLI and semantic phenomena.} NLI has served as a central evaluation task for language understanding \citep{bowman2015snli,williams2018multinli}, with natural logic providing a formal account of the underlying entailment relations \citep{maccartney2009natural} and analyses revealing model sensitivity to annotation artefacts \citep{gururangan2018annotation}, syntactic heuristics \citep{mccoy2019hans}, and specific phenomena such as negation and lexical entailment \citep{geiger2020neural,nie2020anli}. These works often operate at the level of observable outputs; our contribution is orthogonal, investigating the internal representational geometry of the semantic operations that produce NLI predictions.

\paragraph{Representation geometry.} Linear structure in embeddings \citep{mikolov2013distributed} has motivated a line of work showing that entity relations \citep{hernandez2024linearity}, in-context learning tasks \citep{todd2024function}, and high-level semantic properties \citep{park2024linear,marks2024geometry} occupy consistent directions in transformer representation space. We extend this line from individual relations or binary properties to \emph{operation-level} transformations between sentence pairs, providing causal evidence that the identified geometric structure is behaviourally relevant.

\section{Methodology}
\label{sec:method}

\subsection{Setup and Notation}\label{sec:method_notation}
Let $\mathcal{X}$ denote the space of NLI inputs, where each $x=(p,h){\in}\mathcal{X}$ is a premise-hypothesis pair, and let $f_\theta$ be a pretrained transformer with $L$ layers. At each layer $l$, we extract final-token activations from three components: the residual stream $\mathbf{h}^{(l)}_{\texttt{resid}}(x){\in}\mathbb{R}^d$, the attention output $\mathbf{h}^{(l)}_{\texttt{attn}}(x)$, and the MLP output $\mathbf{h}^{(l)}_{\texttt{mlp}}(x)$, following the standard transformer decomposition~\citep{elhage2021mathematical}. We study six semantic operations $o\in\mathcal{O}$: {Negation}, {Antonymy}, {Quantification}, {Abstraction}, {Specification}, and {Co-hyponymy}; definitions and examples are given in Table~\ref{tab:operations} and their formal grounding in compositional semantics and natural logic is provided in App.~\ref{sec:appendix_foundations}. Each operation defines a transformation function $T_o:\mathcal{X}\to\mathcal{X}$ that modifies the hypothesis of an input $x_i$, yielding a contrastive pair $(x_i,T_o(x_i))$. We assume that $T_o$ changes only the intended semantic dimension of the hypothesis; this paired-isolation assumption is enforced through controlled data generation (\S\ref{sec:experimental_setup}), with operation selection details in App.~\ref{sec:appendix_operation_selection}.

\begin{table*}[ht]
\centering
\caption{\textbf{Semantic operations studied,} with linguistic characterisation and representative example pairs. For each operation, a shared premise $P$ is fixed and the hypothesis $H$ is modified: $H_{\mathrm{clean}}$ is entailed by $P$, and $H_o$ denotes the hypothesis after applying operation $o$. The resulting label changes from \textsc{entailment} to \textsc{contradiction} ({Negation}, {Antonymy}, {Quantification}) or \textsc{neutral} ({Abstraction}, {Specification}, {Co-hyponymy}).}
\label{tab:operations}
\small
\begin{tabular}{llp{6.2cm}}
\toprule
\textbf{Operation} & \textbf{Characterisation} & \textbf{Example} \\
\midrule
\multicolumn{3}{l}{\textit{Shared premise:} $P$ = \textit{The company hired three new engineers last month.}} \\
\multicolumn{3}{l}{\textit{Clean hypothesis:} $H_{\mathrm{clean}}$ = \textit{The company hired new engineers.} \quad (\textsc{entailment})} \\
\midrule
{Negation}       & Logical reversal via negation marker          & $H_o$: \textit{The company did not hire new engineers.} \\
{Antonymy}       & Lexical substitution with antonym             & $H_o$: \textit{The company fired new engineers.} \\
{Abstraction}    & Shift from specific to general term           & $H_o$: \textit{The company hired new staff.} \\
{Specification}  & Shift from general to specific term           & $H_o$: \textit{The company hired new software engineers.} \\
{Quantification} & Modification of quantifier scope or strength  & $H_o$: \textit{The company hired no new engineers.} \\
{Co-hyponymy}    & Substitution with co-hyponym (same hypernym)  & $H_o$: \textit{The company hired new designers.} \\
\bottomrule
\end{tabular}
\end{table*}

\subsection{Transformation Effects and Subspace Estimation}
\label{sec:method_subspaces}
Let the activation delta induced by applying operation $o$ to input $x_i$ at layer $l$ (Fig.~\ref{fig:methodology_overview}, \textbf{RQ1}-\textbf{RQ3}) be:
\begin{equation}
\label{eq:delta}
    \Delta_o^{(l)}(x_i) = \mathbf{h}^{(l)}\!\left(T_o(x_i)\right) - \mathbf{h}^{(l)}(x_i)
\end{equation}
Collecting $N$ deltas yields $\mathbf{D}_o^{(l)} \in \mathbb{R}^{N \times d}$. Following \citet{zhao_when_2025}, operation-specific structure is identified via truncated singular value decomposition of the mean-centred matrix $\tilde{\mathbf{D}}_o^{(l)} = \mathbf{D}_o^{(l)} - \bar{\boldsymbol{\delta}}_o^{(l)}$, where $\bar{\boldsymbol{\delta}}_o^{(l)}$ is the column-wise mean of $\mathbf{D}_o^{(l)}$:

\begin{equation}
\label{eq:svd}
    \tilde{\mathbf{D}}_o^{(l)} = \mathbf{U}\,\boldsymbol{\Sigma}\,\mathbf{V}^\top
\end{equation}
The top-$k$ right singular vectors define an orthonormal basis $\mathbf{B}_o^{(l)} \in \mathbb{R}^{k \times d}$ (row-stacked) for the operation subspace $\mathcal{S}_o^{(l)} \subset \mathbb{R}^d$. For any $\boldsymbol{\delta} \in \mathbb{R}^d$, the orthogonal projection onto $\mathcal{S}_o^{(l)}$ is:
\begin{equation}
\label{eq:projection}
\operatorname{proj}_{\mathcal{S}_o^{(l)}}(\boldsymbol{\delta}) = {\mathbf{B}_o^{(l)}}^\top \mathbf{B}_o^{(l)}\,\boldsymbol{\delta}
\end{equation}
This subspace view is a local approximation: applying $T_o$ is modelled through the activation difference it induces at layer $l$, without assuming that semantic transformations act as globally linear maps. The approximation is motivated by evidence that some relational and truth-conditional properties are linearly recoverable in transformer activations \citep{park2024linear,hernandez2024linearity,marks2024geometry}, and by activation-steering work showing that contrastive activation differences can yield behaviourally effective directions \citep{turner2023activation,rimsky2024steering}. We therefore use SVD to estimate the dominant shared directions of $\Delta_o^{(l)}$, interpreting $\mathcal{S}_o^{(l)}$ as the layer-local geometric structure through which operation $o$ is expressed; assumptions and failure modes are discussed in App.~\ref{sec:appendix_geometric_assumptions}.
\subsection{Representational Structure}
\label{sec:method_structure}

The extent to which transformation effects induce geometrically distinct structure in representation space (\textbf{RQ1}) is evaluated using two complementary measures.

\paragraph{Selectivity.} For each subspace $\mathcal{S}_o^{(l)} $, activation deltas from all operations are projected into $\mathcal{S}_o^{(l)}$ and the selectivity ratio is computed:

\begin{equation}
\label{eq:selectivity}
\rho_o^{(l)} =
\frac{
  \mathbb{E}_{x \sim o}\!\left[\left\|\operatorname{proj}_{\mathcal{S}_o^{(l)}}\!\bigl(\Delta_o^{(l)}(x)\bigr)\right\|_2^{2}\right]
}{
  \mathbb{E}_{x \sim \bigcup_{o' \neq o} o'}\!\left[\left\|\operatorname{proj}_{\mathcal{S}_o^{(l)}}\!\bigl(\Delta(x)\bigr)\right\|_2^{2}\right]
}
\end{equation}
A ratio $\rho \gg 1$ indicates that on-target deltas project more strongly onto $\mathcal{S}_o^{(l)}$ than off-target deltas do, indicating a dedicated subspace. Statistical significance is assessed via a one-sided permutation test; full details are in App.~\ref{sec:appendix_metrics}.
\paragraph{Classification.} $\Delta_o$ are projected into operation-specific subspaces and used to train a 6-way logistic regression classifier with 5-fold stratified cross-validation and random-subspace controls. Silhouette scores on t-SNE projections \citep{van2008visualizing} are reported only as a supplementary diagnostic.

\subsection{Causal Intervention} \label{sec:method_steering}
To evaluate whether operation subspaces are causally implicated in prediction (\textbf{RQ2}) and to characterise forward and inverse intervention geometry, targeted activation steering is applied (Fig.~\ref{fig:methodology_overview}, \textbf{RQ2}).
Two direction types are derived from $\Delta_o^{(l)}$: (i) the \emph{principal variance direction}, the top right singular vector of $\mathbf{D}_o^{(l)}$; and (ii) the \emph{mean-shift direction}: 
\begin{equation}
\mathbf{v}_o^{(l)} = \frac{\mathbb{E}_i[\Delta_o^{(l)}(x_i)]}{\left\|\mathbb{E}_i[\Delta_o^{(l)}(x_i)]\right\|_2}
\end{equation}
The latter corresponds to contrastive activation addition (CAA) \citep{turner2023activation,rimsky2024steering}. The residual stream is modified as:
\begin{equation}
\label{eq:steering}
\mathbf{h}^{(l)}(x) \leftarrow \mathbf{h}^{(l)}(x) + \alpha\,\mathbf{v}_o^{(l)}
\end{equation}
In the \emph{forward regime} ($+\alpha$), $\mathbf{v}_o^{(l)}$ is applied to clean inputs to shift predictions toward the label associated with $T_o$. In the \emph{inverse regime} ($-\alpha$), it is applied to transformed inputs to restore the clean prediction. The primary metric is the \emph{flip-to-target rate}: the fraction of examples for which the predicted label changes to the target label induced by $T_o$. Evaluation is conducted at $\alpha \in \{5, 10, 20\}$ across all layers, with the best-performing $\alpha$ selected per layer. Results are compared against random steering directions matched in norm.


\subsection{Mechanistic Localisation}
\label{sec:method_localisation}

To identify where transformation structure emerges and influences prediction (\textbf{RQ4}), two attribution analyses are applied.

\paragraph{Layer-wise logit decoding.} At each layer $l$, the residual-stream representation is decoded using the unembedding matrix $W_u$ \citep{belrose2023eliciting}:
\begin{equation}
\label{eq:logitlens}
\operatorname{logit\_lens}(l) = W_u \cdot \mathbf{h}^{(l)}_{\texttt{resid}}
\end{equation}
The logit difference $\operatorname{logit}(\text{target}) - \operatorname{logit}(\text{corrupt})$ is tracked across layers to identify the \emph{commitment layer}: the depth at which the correct NLI label first consistently dominates.

\paragraph{Direct logit attribution (DLA).} The final logit difference is decomposed into additive contributions from model components \citep{elhage2021mathematical, nostalgebraist2020interpreting}:
\begin{equation}
\label{eq:dla}
\begin{aligned}
\operatorname{logit}(\text{target}) - \operatorname{logit}(\text{corrupt})
=\\ \sum_{l=1}^{L}\Bigl(W_u \cdot \mathbf{h}_{\texttt{attn}}^{(l)} + W_u \cdot \mathbf{h}_{\texttt{mlp}}^{(l)}\Bigr)
\end{aligned}
\end{equation}
This decomposition identifies which attention and MLP layers contribute most strongly to the NLI decision, and whether these align with layers exhibiting high subspace selectivity.

\section{Experimental Setup}\label{sec:experimental_setup}

\paragraph{Models.}
We evaluate four pretrained decoder-only transformers spanning three model families and the 1.7-4B parameter scale: Qwen3-1.7B~\citep{qwen3technicalreport}, Phi-3-mini-4k-instruct~\citep{abdin2024phi3}, Qwen2.5-3B-Instruct~\citep{qwen2.5}, and Gemma3-4B-Instruct~\citep{gemmateam2025gemma3} (hereafter Qwen-1.7B, Phi-3, Qwen-3B, and Gemma-4B). No task-specific fine-tuning is applied. Model selection criteria and architectural details are provided in App.~\ref{sec:model_params}.

\paragraph{Data.}
For each operation in $\mathcal{O}$, we generate $300$ paired NLI examples $(x_i, T_o(x_i))$ using Claude Sonnet 4.5 \citep{anthropic2025claude} with structured prompts, operation-specific constraints, and embedded validation checks. Each example starts from a clean premise-hypothesis; the hypothesis is then modified by a target operation while the premise is unchanged. We enforce paired isolation by requiring the modified hypothesis to change only the intended semantic relation, remain grammatical and plausible, and receive the operation-specific target label. Generation was performed across six thematic domains, producing $1{,}800$ contrastive pairs in total. Full generation and validation criteria are provided in App.~\ref{sec:appendix_data_generation}.

\paragraph{Activation extraction and scoring.}
The component activations in \S\ref{sec:method_notation} correspond to the TransformerLens hook points \texttt{resid\_post}, \texttt{attn\_out}, and \texttt{mlp\_out}. Model predictions are obtained by scoring next-token probabilities over the NLI label tokens and selecting the highest-probability token. 

\paragraph{Hyperparameters and controls.}
The subspace rank is fixed at $k{=}4$, steering interventions are evaluated at $\alpha {\in} \{5, 10, 20\}$; preliminary checks showed qualitatively similar trends for $k {\in} \{2, 8\}$,  and classification uses 5-fold stratified cross-validation with disjoint estimation and evaluation splits. Subspace, steering, and classification metrics are compared against matched random-baseline controls. Random subspace controls use rank-$k$ orthonormal bases sampled via QR orthogonalisation of Gaussian matrices, and random steering controls use norm-matched random directions. Additional implementation details and sensitivity checks are provided in App.~\ref{sec:appendix_controls}.
\section{Results and Analysis}\label{sec:results}
\begin{figure}[t] 
\centering 
\begin{subfigure}[t]{\linewidth} 
\includegraphics[width=\linewidth]{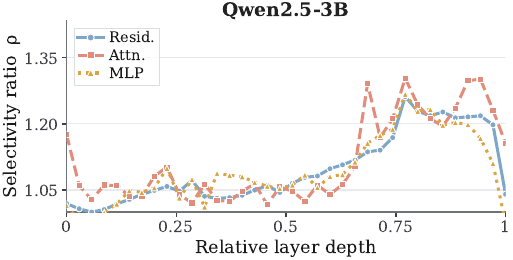} 
\end{subfigure} \hfill 
\begin{subfigure}[t]{\linewidth}
\includegraphics[width=\linewidth]{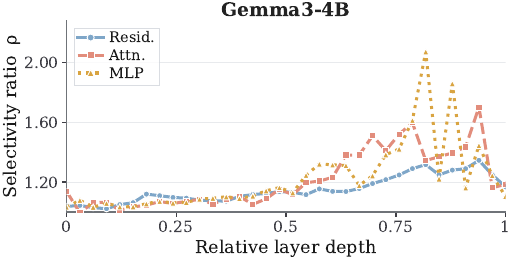} 
\end{subfigure} 
\caption{Layer-wise selectivity ratio $\rho$ across relative depth for Qwen-3B and Gemma-4B. Selectivity is low in early layers ($\rho{\approx}1$) and increases toward the upper network. Full four-model results are in App.~\ref{app:selectivity_curves_full}.} 
\label{fig:fig1_selectivity_curves} 
\end{figure}

\paragraph{RQ1 (Structure): \textbf{Transformation-level effects form partially separable subspaces.}} If models distinguish between operations that produce the same entailment label, operation-specific activation changes should exhibit geometric consistency and separability. We first evaluate this structure using subspace selectivity. Across all model-component configurations, best-layer selectivity is significant under permutation testing, with ratios ranging from $1.16$ to $2.07\times$ ($p{\leq}0.009$; Tab.~\ref{tab:selectivity}). Selectivity is weak in early layers and increases toward the upper network, peaking at relative depths of $0.59$-$0.97$ (Fig.~\ref{fig:fig1_selectivity_curves}). MLP outputs show the strongest selectivity in Phi-3 and Gemma-4B, while attention outputs lead in the Qwen models; the residual stream is weakest across models.

\begin{table}[t]
\centering
\small
\caption{\textbf{Peak (best-layer) subspace selectivity per model and component. $\rho$: selectivity ratio.} $d$: Cohen's $d$. $p$: one-sided permutation $p$-value (null distribution constructed from 1000 random orthonormal subspaces). Rel.\ depth: relative layer position of the peak (0{=}input, 1{=}final layer).}
\label{tab:selectivity}
\resizebox{.5\textwidth}{!}{
\begin{tabular}{llcccr}
\toprule
\textbf{Model} & \textbf{Component} & $\boldsymbol{\rho}$ & $\boldsymbol{d}$ & $\boldsymbol{p}$ & \textbf{Rel.\ depth} \\
\midrule
\multirow{3}{*}{Qwen-1.7B}
  & resid & 1.34 & 0.54 & 0.002 & 0.93 \\
  & attn  & 1.51 & 0.66 & 0.001 & 0.67 \\
  & mlp   & 1.41 & 0.61 & 0.002 & 0.59 \\
\midrule
\multirow{3}{*}{Phi-3}
  & resid & 1.16 & 0.35 & 0.002 & 0.97 \\
  & attn   & 1.52 & 1.11 & 0.001 & 0.90 \\
  & mlp    & 1.90 & 1.23 & 0.005 & 0.77 \\
\midrule
\multirow{3}{*}{Qwen-3B}
  & resid & 1.26 & 0.46 & 0.003 & 0.77 \\
  & attn   & 1.30 & 0.57 & 0.001 & 0.69 \\
  & mlp    & 1.27 & 0.45 & 0.004 & 0.77 \\
\midrule
\multirow{3}{*}{Gemma-4B}
  & resid & 1.35 & 1.09 & 0.001 & 0.94 \\
  & attn & 1.70 & 2.15 & 0.001 & 0.94 \\
  & mlp   & 2.07 & 2.40 & 0.002 & 0.82 \\
\bottomrule
\end{tabular}
}
\end{table}

We then evaluate linear separability after projecting deltas into the learned operation subspaces. Six-way classification accuracy reaches $84.8$-$99.0\%$ at the best layer per model-component combination, substantially above random-subspace baselines (Tab.~\ref{tab:classification}). However, the geometry is not cleanly modular. Per-operation selectivity is heterogeneous: Abstraction has the highest mean selectivity across model-component configurations ($\bar{\rho}{=}2.13$), followed by Quantification ($\bar{\rho}{=}1.96$), Co-hyponymy ($\bar{\rho}{=}1.88$), and Specification ($\bar{\rho}{=}1.21$). Negation and Antonymy fall below $\rho{=}1$ in several configurations, indicating overlap with other operation subspaces. This suggests that label-aligned structure partly interacts with operation-specific information; \textit{transformation effects are not reducible to label-level information, but neither are they fully independent operation modules.} Full per-model classification results are reported in App.~\ref{app:classification_results}; per-operation selectivity results are in App.~\ref{app:full-results}.

\paragraph{RQ2 (Causality): \textbf{Transformation-derived directions can be causally active, but geometric structure does not guarantee steerability.}} We investigate whether transformation-induced directions affect predictions under activation steering, comparing CAA and SVD directions (\S\ref{sec:method_steering}). Activation steering changes predictions in three of the four models, indicating that transformation-level directions can have causal influence on model outputs. However, this causal effect is not directly predicted by geometric selectivity: Gemma-4B has the strongest peak selectivity in RQ1 ($\rho{=}2.07$), but exhibits near-zero flip-to-target rates under both CAA and SVD steering. As RQ4 shows, Gemma-4B also exhibits an MLP-dominant, three-phase logit trajectory, suggesting that its operation-level structure may be geometrically organised but not easily controlled by single-direction residual-stream steering. Thus, \textit{geometric separability and causal accessibility should be treated as distinct properties.}

\begin{table}[t]
\centering
\small
\caption{CAA vs SVD steering comparison. Each row pools over operations within a model and direction regime. \emph{Win rate}: proportion of operations for which CAA achieves higher flip-to-target rate than SVD. $\bar{\Delta}$: mean pair difference in flip-to-target rate (positive values favour CAA). $p$: one-sided paired permutation $p$-value (10{,}000 permutations; App.~\ref{sec:app_statistical_testing}). The pooled row combines all three models and both regimes.}
\label{tab:steering_summary}
\resizebox{.5\textwidth}{!}{
\begin{tabular}{llccc}
\toprule
\textbf{Model} & \textbf{Regime} & \textbf{Win rate} & $\boldsymbol{\bar{\Delta}}$ &
  $\boldsymbol{p}$ \\
\midrule
\multirow{2}{*}{Qwen-1.7B}
  & forward & 0.667 & +0.160 & 0.062 \\
  & inverse & 0.000 & $-$0.337 & 1.000 \\
\midrule
\multirow{2}{*}{Qwen-3B}
  & forward & 0.500 & +0.113 & 0.304 \\
  & inverse & 0.500 & +0.023 & 0.424 \\
\midrule
\multirow{2}{*}{Phi-3}
  & forward & 0.833 & +0.210 & 0.062 \\
  & inverse & 0.833 & +0.223 & 0.033 \\
\midrule
\multicolumn{2}{l}{\textbf{Pooled (3 models, both regimes)}} & \textbf{0.556} &
  \textbf{+0.066} & \textbf{0.157} \\
\bottomrule
\end{tabular}
}
\end{table}

\textit{The divergence between CAA and SVD directions reinforces this point.} Their cosine similarity varies across operations, from $0.459$ for Quantification to $0.810$ for Specification, suggesting that they capture different aspects of the transformation effect. CAA outperforms SVD in $20$ of $33$ non-tied comparisons, with a non-tied win rate of $0.606$ and a pooled win rate of $0.556$ when ties are included (Tab.~\ref{tab:steering_summary}). This advantage is not uniform across models or regimes: Phi-3 favours CAA in both regimes, Qwen-3B is balanced, and Qwen-1.7B reverses across regimes, with CAA leading in the forward setting but SVD outperforming CAA on all six operations in the inverse setting. At the pooled level, the CAA advantage is not statistically significant ($\bar{\Delta}{=}0.066$, $p{=}0.157$). A sensitivity analysis at $\alpha \in \{0.5,1,2\}$ shows that steering effects decrease substantially at smaller intervention magnitudes (App.~\ref{sec:app_steering_per_direction}).

\begin{figure}[ht]
    \centering
    \includegraphics[width=\linewidth]{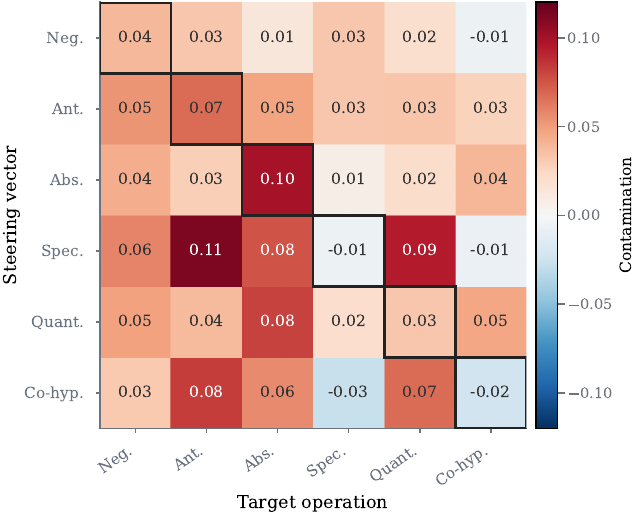}
    \caption{Cross-operation steering contamination. Each cell gives the excess flip-to-target rate when an SVD direction derived from a source operation (rows) is applied to inputs from a target operation (columns), relative to a matched random-direction baseline, averaged over models and layers at $\alpha{=}20$. On-diagonal cells index within-operation steering; off-diagonal cells index cross-operation interference.}
    \label{fig:fig3_cross_op_heatmap}
\end{figure}

\paragraph{RQ3 (Dissociation): \textbf{Cross-operation interference is sparse but structured, and selectivity does not imply independence.}} We measure cross-operation interference by applying an SVD direction derived from one source operation to examples from each target operation. This tests whether steering effects remain confined to the source operation or transfer to other transformations. Off-target effects are generally sparse: at $\alpha{=}20$, the off-diagonal contamination values are right-skewed, with most pairs clustered near zero (Fig.~\ref{fig:fig3_cross_op_heatmap}). The five largest off-diagonal values are Specification~$\to$~Antonymy ($0.112$), Specification~$\to$~Quantification ($0.095$), Co-hyponymy~$\to$~Antonymy ($0.083$), Quantification~$\to$~Abstraction ($0.081$), and Specification~$\to$~Abstraction ($0.076$); all other off-diagonal pairs are at most $0.07$. Thus, interference is not uniformly distributed, but concentrated among a small number of semantically adjacent or label-related operation pairs. Per-model and inverse-regime results are reported in App.~\ref{sec:app_cross_op}.

The dissociation between selectivity and independence runs in both directions. Specification has moderate mean selectivity ($\bar{\rho}{=}1.21$) but is the strongest source of cross-operation interference: Specification~$\to$~Antonymy ($0.112$) and Specification~$\to$~Quantification ($0.095$) are the two largest off-diagonal values, and Specification appears as source in three of the five largest pairs. Antonymy shows a different pattern: althoguh it is not the main source of interference,  it is frequently affected by other operations, with two of the five largest off-diagonal values targeting it. This pattern is consistent with its low subspace selectivity and overlap with other operation subspaces. Abstraction, which has the highest selectivity ($\bar{\rho}{=}2.13$), has a substantially lower source-side mean ($0.028$). Negation has both weak selectivity ($\bar{\rho}{=}1.03$) and the lowest source-side interference ($0.018$), consistent with a diffuse rather than cleanly isolated signal. Per-model patterns are in App.~\ref{sec:app_cross_op}; interference is broadest for Qwen-3B and most concentrated for Qwen-1.7B, while Gemma-4B shows near-zero contamination. The results indicate that \textit{subspace concentration is neither necessary nor sufficient for intervention-level independence; separability, selectivity, source-side interference, and target-side susceptibility should be treated as distinct mechanistic properties.}

\begin{figure}[t] 
\centering 
    \begin{subfigure}[t]{\linewidth}        
    \includegraphics[width=\linewidth]{ 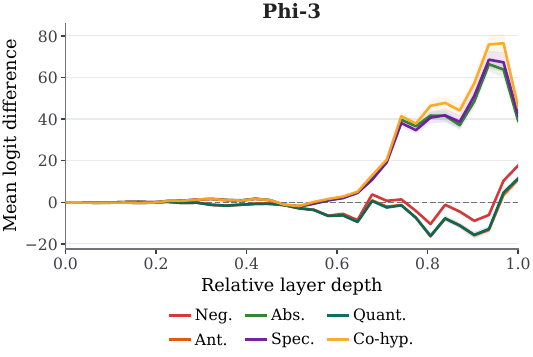} 
    \end{subfigure} 
    \hfill 
    \begin{subfigure}[t]{\linewidth} 
    \includegraphics[width=\linewidth]{ 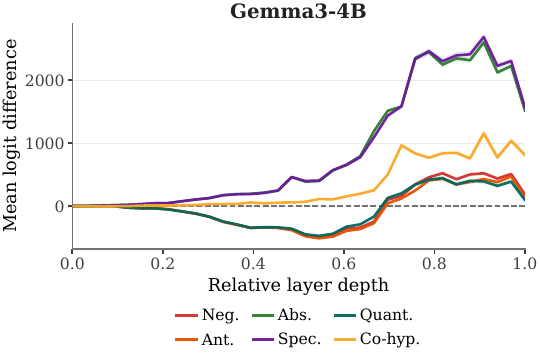} 
    \end{subfigure} 
    \caption{Sample layer-wise logit difference trajectories for Phi-3 and Gemma-4B. Phi-3 shows near-zero evidence until relative depth $0.6$-$0.7$, a brief suppression phase around $0.70$-$0.75$, then sharp rise. Gemma-4B exhibits a three-phase pattern: gradual early rise, mid-network suppression, and strong late amplification with substantially larger magnitudes. 
    }
    \label{fig:rq4_logit_lens_trajectories} 
\end{figure}

\paragraph{RQ4 (Localisation): \textbf{Transformation-dependent label evidence is amplified in  mid-to-late layers, with architecture-dependent component pathways.}} Layer-wise logit decoding shows that label evidence generally emerges in mid-to-late layers and is amplified in the upper network (Fig.~\ref{fig:rq4_logit_lens_trajectories}; App.~\ref{app:logit_lens_full}). Qwen-1.7B and Phi-3 remain near zero until relative depth $0.6$-$0.7$, while Qwen-3B aligns later, rising sharply in the final quarter of the network. Across models, peak logit differences occur at relative depths of $0.86$-$0.94$. Gemma-4B follows a distinct three-phase trajectory: gradual early rise, mid-network suppression around relative depth $0.6$-$0.65$, and strong late amplification at $0.75$-$0.85$.

Direct logit attribution shows that component pathways vary by model family (Fig.~\ref{fig:rq4_dla_component_balance}; App.~\ref{app:dla}). Phi-3 is primarily attention-driven, Gemma-4B is MLP-dominant for most operations, Qwen-1.7B is operation-dependent, and Qwen-3B is comparatively balanced. No consistent ordering is observed in the depth at which operations first favour the target label. \textit{These results identify a shared representational phenomenon, but not a shared circuit: transformation-level structure appears across models, while its computational implementation remains architecture-dependent.}

\paragraph{Overall Interpretation.}
The results indicate that transformation-level effects are geometrically organised in model representations, but neither label-independent nor operation-modular. Models do not appear to encode only the final NLI label: transformation effects are linearly decodable, show significant subspace selectivity, and can influence predictions under steering. At the same time, these operations are not represented as cleanly separate modules: their subspaces overlap, steering effects can transfer across operations, and causal accessibility varies across models and regimes. Transformation-level effects are therefore structured, partially shared, and architecture-dependent. Label-level analyses undercharacterise this structure, while stronger circuit-level claims require further localisation beyond the subspace and intervention evidence provided here.

\section{Conclusion}\label{sec:conclusion}
This paper investigated whether transformer models encode transformation-level effects as structured geometric objects in activation space, distinct from the NLI labels they produce. Using controlled contrastive premise-hypothesis pairs, we measured transformation-induced activation differences, analysed their subspace structure, and tested their causal effect through activation steering.

The results indicate that transformation-level effects are geometrically organised in model representations, but neither label-independent nor operation-modular. Models do not appear to encode only the final NLI label: transformation effects are linearly decodable, show significant subspace selectivity, and can influence predictions under steering. At the same time, these operations are not represented as cleanly separate modules: their subspaces overlap, steering effects can transfer across operations, and causal accessibility varies across models and regimes. Transformation-level effects are therefore structured, partially shared, and architecture-dependent.

These findings constrain mechanistic accounts of semantic inference in transformer models. Label-level analyses undercharacterise this structure, while stronger circuit-level claims require further localisation beyond the subspace and intervention evidence provided here.


\section*{Limitations}\label{sec:limitations}
This analysis is restricted to decoder-only models at the 1.7-4B scale; generalisation to other architectures, larger models, or explicitly supervised NLI systems remains an open question. The paired-isolation assumption is enforced through controlled data construction but cannot be perfectly realised; residual confounds may remain despite random-baseline controls. For example, operations are partially aligned with NLI labels in the controlled dataset, which creates a genuine identifiability concern. We address this with within-label prediction, label-subspace residualisation, and label-aware contamination controls (App.~\ref{app:label_controlled_operations}), but fully crossing every operation with every NLI label remains future work. The mechanistic characterisations of semantic operations could in principle be used to construct adversarial inputs that exploit operation-specific directions. Finally, the framework identifies geometric structure and causal relevance but does not localise the underlying circuits. Connecting subspace-level findings to specific mechanistic components remains an important direction for future work.

\section*{Ethical considerations} 
This work investigates whether semantic operations in LLMs are represented as structured and causally active geometric effects in activation space. Improving mechanistic understanding of semantic inference may support transparency, interpretability, and safety research by enabling more precise analysis of how models process data and produce inference decisions. On the other hand, identifying operation-specific directions and steerable subspaces introduces potential misuse risks. For instance, representations could be exploited to design targeted activation interventions, adversarial prompts, or inference manipulations that selectively alter model behaviour while remaining difficult to detect through output-level analysis.

\bibliography{custom}

\appendix
\section{Linguistic and Logical Foundations}
\label{sec:appendix_foundations}

\subsection{Sentence Meaning as Functional Constraint}
Formal semantics identifies the meaning of a sentence with a function of the meanings of its parts~\citep{Frege1892}. Under this compositional view, $\llbracket S \rrbracket$ is determined by the meanings of $S$'s constituent expressions $s_1, s_2, \ldots, s_n$ with their syntactic combination~\citep{Montague1970a,Montague1970b}: 
\begin{equation}
\llbracket S \rrbracket \;=\; f\!\left(
  \llbracket s_1 \rrbracket,\;
  \llbracket s_2 \rrbracket,\;
  \ldots,\;
  \llbracket s_n \rrbracket
\right),
\label{eq:compositionality}
\end{equation}
so that to know what $S$ means is to know how its parts contribute to the whole. This determines which worlds $S$ permits and which it excludes, and sentence-pair relations follow directly: $P$ entails $H$ when $\llbracket P \rrbracket^{-1}(1) \subseteq \llbracket H \rrbracket^{-1}(1)$, $P$ contradicts $H$ when the two sets are disjoint, and the pair is neutral otherwise.

Natural language inference (NLI) provides a task-level approximation to this semantic picture~\citep{bowman2015snli,williams2018multinli}. Given a premise $P$ and a hypothesis $H$, the problem is to determine which of these three relations holds; a model that solves it correctly must therefore be sensitive, at some level, to the compositional meaning constraints that determine the answer. This makes NLI a principled testbed for studying how models represent sentence meaning: not only through their final labels, but through the relation the labels reflect.

\subsection{Semantic Operations as Local Meaning Changes}
\label{app:semantic_operations}
A semantic operation acts locally: it modifies a targeted expression or semantic part within the hypothesis while keeping the premise, and as much of the remaining hypothesis as possible fixed. Under the compositional view of Eq.~\ref{eq:compositionality}, this local change can alter the meaning of the whole sentence and the semantic relation between premise and hypothesis. This link between constituent-level change and sentence-level consequence makes operations an appropriate unit of analysis for studying \textit{how meaning changes are encoded in model representations}.

Natural logic formalises this idea through a small set of semantic relations between expressions: forward entailment ($\sqsubset$), reverse entailment ($\sqsupset$), equivalence ($\equiv$), negation ($\wedge$), alternation ($\mid$), cover ($\smile$), and independence ($\#$)~\citep{maccartney2009natural}. These local relations interact with the related context, where that changing a word, an operator, or the level of specificity can alter the relation between the full premise and the hypothesis. For example, replacing a term with a more specific expression makes the hypothesis stronger than the premise supports ($\sqsubset$). In each case, the edit is local but its inferential effect is sentence-wide.

The six operations studied in this paper instantiate this general pattern across several relation types: {Negation} and {Quantification} modify logical structure; {Antonymy} and {Co-hyponymy} modify lexical relations; and {Abstraction} and {Specification} modify semantic granularity. They provide a controlled set of meaning-altering transformations, allowing us to ask whether models distinguish the operation that changes the relation from the label it produces.

\subsection{Atomicity and Paired Isolation}
\label{app:atomicity}

The operations defined above are local by design, but locality alone does not guarantee that an activation difference reflects only the intended semantic change. We use \emph{atomicity} in an operational sense: within a constructed pair, the transformed hypothesis should differ from the clean hypothesis along one primary semantic dimension. A \op{Specification} pair, for instance, should add a more specific constraint without introducing an unrelated contradiction.

This motivates \emph{paired isolation}. Each example consists of a clean input $(P, H_{\text{clean}})$ and a transformed input $(P, H_o)$, where the premise is fixed and the hypothesis is modified by a target operation $o$. Under atomicity, the activation difference can be defined as:
\begin{equation}
\small
\Delta^{(l)}_o(x_i)
\;=\;
\underbrace{\boldsymbol{\delta}^{(l)}_o}_{\text{operation effect}} +
\underbrace{\boldsymbol{\varepsilon}^{(l)}_i}_{\text{instance noise}},
\qquad
\mathbb{E}_i\!\left[\boldsymbol{\varepsilon}^{(l)}_i\right] \approx \mathbf{0},
\label{eq:atomic_decomp}
\end{equation}
where $\boldsymbol{\delta}^{(l)}_o$ is the shared operation-induced activation difference and $\boldsymbol{\varepsilon}^{(l)}_i$ is zero-mean instance noise. The activation difference is intended to capture the representational change associated with $T_o$, rather than random variation between unrelated parts.

The assumption is approximate. A negation marker may also alter syntactic structure; an antonym may differ from the original predicate in frequency or register; and a specification may introduce additional lexical material. The role of controlled generation and validation is to make the target operation the dominant source of variation in $\Delta^{(l)}_o$, not to guarantee a perfectly isolated semantic atom. The main analyses combine several checks: random-subspace controls test whether the observed structure exceeds arbitrary low-dimensional projections; label-controlled analyses address the largest systematic confound; and cross-operation contamination tests whether operation-derived directions remain specific or transfer to other transformations.

\paragraph{Connection to the mechanistic analysis.} A model whose internal representations are sensitive to sentence meaning should, under this view, respond systematically when a semantic operation is applied. The compositional structure of Eq.~\ref{eq:compositionality} motivates the central test: whether a local change to the hypothesis, with the premise and surrounding context held fixed, leads to a consistent activation difference in the model's representation of the full input. This activation difference, $\Delta^{(l)}_o(x_i)$, is the quantity our mechanistic analysis is designed to detect. Whether it is structured, operation-specific, and causally active in the model's predictions is the central question the paper inveistage.
\section{Data Generation Details}
\label{sec:appendix_data_generation}
This section describes the generation procedure for the contrastive NLI pairs used in the main experiments. The dataset was an English dataset, generated using Claude Sonnet 4.5 \citep{anthropic2025claude} under a structured prompt with operation-specific constraints. The goal was to produce paired examples in which the clean pair expressed entailment and the modified pair differed from it by exactly one controlled semantic transformation. As the main analysed quantity is $\Delta_o^{(l)} = \mathbf{h}^{(l)}(P, H_o) - \mathbf{h}^{(l)}(P, H_{\text{clean}})$, the dataset must minimise all variation between the clean and modified hypotheses except the target operation; this motivates the controlled design described below. 

\subsection{Operation Selection Criteria} \label{sec:appendix_operation_selection} 
The six operations were selected to provide coverage of the primary lexical semantic relation types identified in natural logic accounts of NLI \citep{maccartney2009natural}, while satisfying the methodological requirements of the mechanistic analysis.

\paragraph{Grounding in natural logic.} 
The six operations instantiate distinct lexical relation types within the natural logic framework of \citet{maccartney2009natural} (see App.~\ref{app:semantic_operations} for the formal grounding). \op{Negation} generates the exhaustive exclusion relation ($\wedge$); \op{Antonymy} and \op{Co-hyponymy} both instantiate alternation ($\mid$) but are distinguished here by structural and representational criteria: antonymy involves direct semantic opposition, whereas co-hyponymy involves category sibling substitution under a shared hypernym. \op{Quantification} generates exclusion via quantifier substitution (e.g.\ \emph{some} $\to$ \emph{no}). \op{Abstraction} and \op{Specification} instantiate reverse and forward entailment ($\sqsupset$, $\sqsubset$) via hypernymy-axis substitution, producing \textsc{neutral} labels.

\paragraph{Selection criteria.} Three criteria governed the final selection. First, each operation must modify the hypothesis along a single semantic dimension, satisfying the paired-isolation requirement of the mechanistic analysis (\S\ref{sec:method_notation}). This excludes operations requiring simultaneous modification of multiple constituents or world-knowledge dependencies that cannot be controlled lexically. Second, the set must cover both non-entailment label outcomes (\textsc{contradiction} and \textsc{neutral}), which is necessary for the label-controlled analyses in Appendix~\ref{app:label_controlled_operations}. Third, operations should be formally and structurally distinct enough to test whether transformation-level structure generalises across qualitatively different types of meaning-altering edits.

\paragraph{Excluded operations.} Meronymy (part-whole substitution) was piloted but produced inconsistent label assignments due to pragmatic variability in part-whole entailment relations, violating the first criterion. Scalar implicature was excluded because label assignment depends on pragmatic context rather than lexical substitution alone. Presupposition-altering operations and causal or temporal modifications were excluded because they require multi-constituent changes or introduce uncontrollable world-knowledge dependencies. The six retained operations represent the maximal set satisfying all three criteria under the controlled generation protocol described in \S\ref{sec:appendix_data_generation_prompting}.

\subsection{Sentence Universe}
\label{sec:appendix_sentence_universe}
\begin{figure*}[t]
\centering
\includegraphics[width=\linewidth]{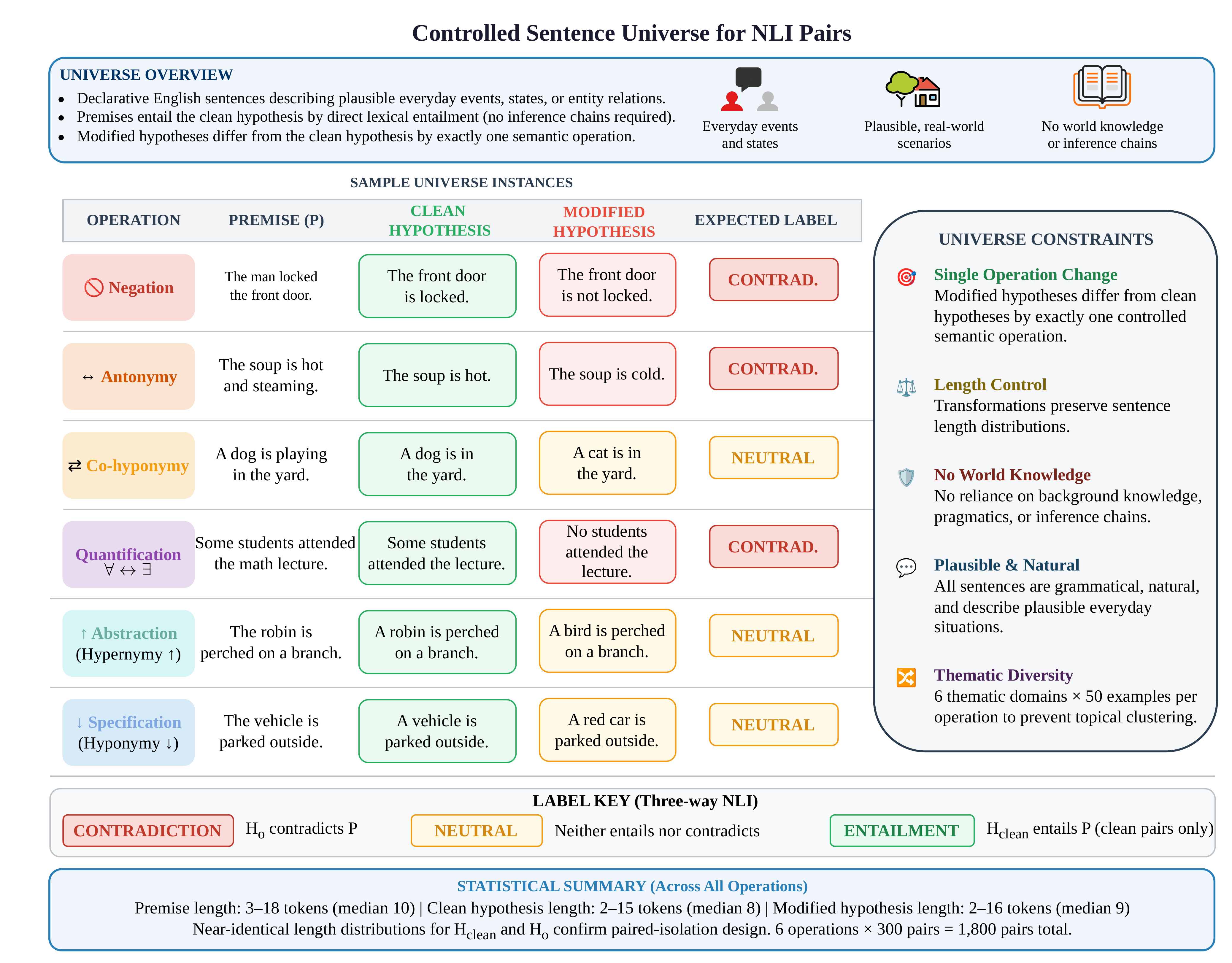}
\caption{Controlled sentence universe for contrastive NLI pair generation. Each pair consists of a fixed premise $P$, a clean hypothesis $H_{\text{clean}}$ (labelled \textsc{entailment}), and a modified hypothesis $H_o = T_o(H_{\text{clean}})$ differing from $H_{\text{clean}}$ by exactly one semantic operation. One representative example is shown per operation.}
\label{fig:data_universe}
\end{figure*}

Figure~\ref{fig:data_universe} provides an overview of the controlled sentence universe. The universe consists of declarative English sentences describing plausible everyday events, states, or entity relations. Premises are designed to have a clean hypothesis by direct lexical entailment, without requiring inference chains, background knowledge, or world-knowledge dependencies. Modified hypotheses differ from the clean hypothesis by a semantic operation.

\paragraph{Sentence length and complexity.}

Table~\ref{tab:dataset_stats} reports descriptive statistics for the generated pairs. Premises range from 7 to 18 tokens (median 10), clean hypotheses from 2 to 15 tokens (median 8), and modified hypotheses from 2 to 16 tokens (median 9) across all operations. The near-identical length distributions of $H_{\text{clean}}$ and $H_o$ confirm that transformations do not systematically alter sentence length, consistent with the paired-isolation design. \op{Quantification} produces the longest sentences (premise median 12, hypothesis median 10--11), reflecting the need to include explicit quantifier expressions. \op{Antonymy} produces the shortest hypotheses (median 4 tokens), consistent with the targeted single-word substitution it requires.

\subsection{Prompting Procedure}
\label{sec:appendix_data_generation_prompting}
The generator model received a structured prompt organised into five components: (i) an operation definition specifying the semantic transformation in formal terms; (ii) a set of word pair examples illustrating valid substitutions for that operation; (iii) explicit label flip rules specifying the required baseline and target labels; (iv) a validation checklist that the model was required to apply to every example before inclusion; and (v) an output format specification requiring structured JSON with fields for premise, clean hypothesis, modified hypothesis, baseline label, target label, operation type, original span, and modified span. Table~\ref{tab:prompt_structure} summarises the prompt components and their functions.\\

\begin{table}[h]
\centering
\small
\caption{Prompt structure used for controlled data generation.}
\label{tab:prompt_structure}
\begin{tabular}{p{0.28\linewidth}p{0.62\linewidth}}
\toprule
\textbf{Component} & \textbf{Function} \\
\midrule
Operation definition  & Defines the semantic edit and target label transition in formal terms. \\
Substitution examples & Provides valid word-pair examples for the target operation. \\
Label flip rules      & Specifies baseline \textsc{entailment} and target label explicitly. \\
Validation checklist  & Lists conditions every example must satisfy before inclusion. \\
Output schema         & Requires structured JSON with fixed fields for downstream filtering. \\
Regeneration rule     & Instructs the model to discard and replace any failing example. \\
\bottomrule
\end{tabular}
\end{table}

\paragraph{Operation definitions in the prompt.}
Each operation was defined in semantic rather than surface-form terms, 
with explicit label flip rules:

\begin{itemize}[leftmargin=*,itemsep=2pt]
    \item \textbf{Negation}: insert a negation marker into the main predicate of the hypothesis; baseline \textsc{entailment} $\to$ target \textsc{contradiction}.
    \item \textbf{Antonymy}: replace a content word with a direct, non-gradable antonym; baseline \textsc{entailment} $\to$ target \textsc{contradiction}.
    \item \textbf{Quantification}: replace a quantifier with a maximally distinct alternative; baseline \textsc{entailment} $\to$ target \textsc{contradiction}.
    \item \textbf{Abstraction}: replace a specific term with a superordinate hypernym; baseline \textsc{entailment} $\to$ target \textsc{neutral}.
    \item \textbf{Specification}: replace a general term with a more specific hyponym; baseline \textsc{entailment} $\to$ target \textsc{neutral}.
    \item \textbf{Co-hyponymy}: replace a term with a sibling category member sharing the same hypernym; baseline \textsc{entailment} $\to$ target \textsc{neutral}.
\end{itemize}

\paragraph{Validation requirements embedded in the prompt.}
The prompt required the generator to verify the following conditions for every example before inclusion:
\begin{enumerate}[leftmargin=*,itemsep=2pt] 
    \item Both clean and modified hypotheses are grammatical and natural English sentences. 
    \item Both describe plausible real-world situations. 
    \item The clean hypothesis is unambiguously entailed by the premise. 
    \item The modified hypothesis bears the target non-entailment label. 
    \item The modified hypothesis does not accidentally remain entailed by the premise. 
    \item The transformation involves exactly one semantic change, with no additional lexical or syntactic modifications if not necessary. 
\end{enumerate}

\paragraph{Self-filtering.} Generation included an explicit self-check stage: the model was instructed to discard and regenerate any example failing any of the above conditions. Low-margin examples, those for which a competent reader might hesitate before assigning the intended label, were identified as a major failure mode in pilot generations and were explicitly prohibited.

\paragraph{Scale and batching.} Each operation was generated in six thematic batches of 50 examples, yielding 300 examples per operation and 1,800 pairs in total across six operations. Each operation was generated independently and saved as a separate file.

\paragraph{Output format.} The prompt required JSON output. A representative example for \op{Negation} is shown below: 
\begin{verbatim}
{
  "id": "neg_0042",
  "premise":   "The company hired three new 
                engineers last month.",
  "clean":     "The company hired new engineers.",
  "corrupted": "The company did not hire 
                new engineers.",
  "baseline_label": "entailment",
  "modified_label": "contradiction",
  "operation":      "NEGATION",
  "original_span":  "hired",
  "modified_span":  "did not hire"
}
\end{verbatim}

\subsection{Filtering and Validation}
\label{sec:appendix_data_generation_filtering} 
After generation, files were inspected programmatically to ensure schema consistency. Each JSON object was validated to contain the required fields: \texttt{id}, \texttt{premise}, \texttt{clean}, \texttt{corrupted}, \texttt{baseline\_label}, \texttt{modified\_label}, \texttt{operation}, \texttt{original\_span}, and \texttt{modified\_span}. 

\subsection{Relation to the Main Experiments}
\label{sec:appendix_data_generation_main}

The purpose of the synthetic generation procedure was not to approximate the full distribution of naturally occurring NLI examples, but to construct minimally varying contrastive pairs suitable for mechanistic analysis. The controlled procedure therefore prioritises paired isolation and label clarity over ecological breadth. 

Since the data are generated by LLMs under explicit structural constraints, residual artefacts may remain, including lexical regularities or templatic biases weaker in naturally occurring corpora. The random-subspace controls, cross-model comparisons, and cross-operation analyses help to bound this risk, but do not eliminate it. The results should therefore be interpreted as evidence about transformation-level structure under controlled contrastive conditions, rather than as a claim about the full natural distribution of NLI examples.

\begin{table}[h]
\centering
\small
\caption{Descriptive statistics of the generated sentence universe. Values reported as median (min--max) per field and operation.}
\label{tab:dataset_stats}
\begin{tabular}{lccc}
\toprule
\textbf{Operation} & \textbf{Premise} & \textbf{$H_{\text{clean}}$} & \textbf{$H_o$} \\
\midrule
\op{Abstraction} & 11 (7--15) & 9 (5--13) & 11 (8--15) \\
\op{Antonymy} & 10 (7--15) & 5 (4--6) & 5 (4--6) \\
\op{Co hyponyms} & 8 (4--13) & 8 (4--12) & 8 (4--13) \\
\op{Negation} & 10 (7--14) & 6 (3--13) & 8 (4--14) \\
\op{Quantification} & 12 (8--18) & 10 (5--15) & 11 (6--16) \\
\op{Specification} & 8 (7--13) & 6 (4--11) & 6 (4--11) \\
\midrule
\textit{All operations} & 10 (4--18) & 8 (4--15) & 9 (4--16) \\
\bottomrule
\end{tabular}
\end{table}

\section{Evaluation Metrics, Controls, and Implementation Details}
\label{sec:appendix_methods}

\subsection{Criteria for Transformation-Based Reasoning}
\label{sec:appendix_criteria}

We operationalise transformation-based reasoning as the existence of structured, reusable directional effects in representation space. A model exhibits transformation-based reasoning with respect to operation $o$ if four criteria are satisfied. \textbf{(i) Consistency}: the transformation induces a stable subspace $\mathcal{S}_o^{(l)}$ that generalises across input instances. \textbf{(ii) Separability}: distinct operations occupy distinguishable regions of representation space, such that a linear classifier separates them substantially above chance. \textbf{(iii) Causality}: the transformation can be induced or reversed via targeted intervention along directions in $\mathcal{S}_o^{(l)}$, producing the corresponding shift in model output. \textbf{(iv) Independence}: intervening along $\mathbf{v}_o$ does not systematically shift model predictions toward the label of a different operation $o'$, as measured by cross-operation steering contamination. Criteria (i)-(ii) are correlational, establishing that structure exists in representations. Criteria (iii)-(iv) are causal, establishing that the structure is used by the model. The main evaluation (\S\ref{sec:method}, \S\ref{sec:results}) tests each criterion via distinct analyses: selectivity and classification address (i)-(ii); activation steering addresses (iii); cross-operation interference addresses (iv).

We analyse the displacement distribution $\Delta_o^{(l)}$ in two complementary ways. The mean-shift direction captures the average operation-induced displacement and is the direction used in contrastive activation addition (CAA)~\citep{turner2023activation,rimsky2024steering}. The SVD basis captures the dominant axes of variation around this displacement distribution. This distinction matters because a semantic operation may be expressed partly as a common shift shared across examples and partly as structured variation due to lexical, syntactic, or contextual realisation. Comparing CAA and SVD tests whether causal control is better explained by the average displacement or by a principal direction of operation-specific variation. 

This formulation connects representation geometry with causal intervention. Prior work shows that relational and truth-conditional information can be recovered from linear structure in transformer activations~\citep{park2024linear,hernandez2024linearity, marks2024geometry}, while activation-steering work shows that contrastive activation differences can change model behaviour when added to internal states~\citep{turner2023activation,rimsky2024steering,zou2023representation}. Related subspace-based interventions have also been used to separate behaviourally relevant representational factors~\citep{zhao_when_2025}. 

The operation subspace $\mathcal{S}_o^{(l)}$ captures the principal directions along which $T_o$ displaces representations at layer $l$, analogous to a tangent space in differential geometry. The analogy is informal: the underlying representation manifold need not be smooth, and $\mathcal{S}_o^{(l)}$ is a linear approximation to a potentially nonlinear displacement structure.

\subsection{Evaluation Metrics}\label{sec:appendix_metrics}
\paragraph{Subspace selectivity.}
\label{sec:app_selectivity}
The selectivity ratio $\rho_o^{(l)}$ (Equation~\ref{eq:selectivity}) measures the relative projection energy of an operation onto its own subspace versus the average projection energy of other operations onto the same subspace. A ratio of $\rho = 1$ indicates no preferential response; values greater than 1 indicate selective response. Formally, for operation $o$ at layer $l$, the selectivity ratio is defined as:
\begin{equation}
\label{eq:selectivity_full}
\rho_o^{(l)} =
\frac{
  \displaystyle\frac{1}{N_o}\sum_{i=1}^{N_o}
    \left\|\operatorname{proj}_{\mathcal{S}_o^{(l)}}\!\bigl(\Delta_o^{(l)}(x_i)\bigr)\right\|_2^{2}
}{
  \displaystyle\frac{1}{\sum_{o' \neq o} N_{o'}}
    \sum_{o' \neq o} \sum_{j=1}^{N_{o'}}
      \left\|\operatorname{proj}_{\mathcal{S}_o^{(l)}}\!\bigl(\Delta_{o'}^{(l)}(x_j)\bigr)\right\|_2^{2}
}
\end{equation}

The numerator is the mean projection magnitude of operation $o$'s deltas onto $\mathcal{S}_o^{(l)}$ (on-target); the denominator pools all deltas from operations $o' \neq o$ onto the same subspace (off-target). Selectivity is reported as a ratio rather than a difference to ensure scale invariance across operations and layers with different overall activation magnitudes.

\paragraph{Cluster Metrics.}
\label{sec:app_cluster}
To assess geometric separation independently of classifier choice, we compute the silhouette score on projected activation deltas:
\begin{equation}
    s(i) = \frac{b(i) - a(i)}{\max\{a(i), b(i)\}},
\end{equation}
where $a(i)$ is the mean intra-cluster distance and $b(i)$ is the mean nearest-cluster distance for sample $i$. Values range from $-1$ to $1$, with higher values indicating better-separated clusters.

\paragraph{Steering and cross-operation contamination.}\label{sec:app_steering}
For causal intervention experiments (\S\ref{sec:method_steering}), we report:
\begin{itemize}[topsep=2pt,itemsep=2pt]
    \item \textbf{Flip-to-target rate}: fraction of examples whose predicted NLI label changes from the pre-intervention label to the target label (the label associated with the applied operation).
    \item \textbf{Contamination/Off-target flip rate}: fraction of examples whose predicted label changes to a label other than the target.
\end{itemize}
The term \emph{contamination}, used in \S\ref{sec:results} and Fig.~\ref{fig:fig3_cross_op_heatmap}, refers to a distinct quantity defined in the cross-operation steering analysis: the excess flip-to-target rate when a source operation's direction is applied to a different target operation, relative to a matched random-direction baseline.

\subsection{Statistical Testing}
\label{sec:app_statistical_testing}

\paragraph{Selectivity significance.}
Statistical significance is assessed using a one-sided permutation test that evaluates whether the observed on-target projection magnitudes exceed those expected under random subspace alignment. For each configuration, an empirical null distribution is constructed by sampling random orthonormal subspaces of rank $k$ and recomputing the selectivity ratio. Reported $p$-values correspond to the proportion of random samples that yield selectivity ratios greater than or equal to the observed value. Cohen's $d$ is reported as an effect size.

\paragraph{Classification significance.}
We compare SVD-based classification accuracy against the random-baseline distribution using permutation tests with 1000 random-subspace draws. We report the proportion of random draws that exceed the observed accuracy as an empirical $p$-value.

\paragraph{Paired permutation test for direction comparison.}
To compare CAA mean-shift and SVD principal variance directions under activation steering (\S\ref{sec:results}, RQ2), we use a paired sign-flip permutation test. Each unit of comparison is a (model, direction, operation) triple for which a flip-to-target rate is available for both direction types, yielding $n = 36$ paired rows (three models, two steering regimes, six operations; Gemma3-4B excluded as non-informative, \S\ref{sec:results}). The test statistic is the mean pair difference $\bar{\Delta} = \mathrm{mean}(\mathrm{CAA} - \mathrm{SVD})$. Under the null hypothesis of no systematic difference, the sign of each pair difference is randomly flipped across $10{,}000$ permutations. The one-sided $p$-value is the proportion of permuted means that equal or exceed the observed $\bar{\Delta}$. 

\subsection{Hyperparameters and Random-Baseline Controls}\label{sec:appendix_controls}
\paragraph{Subspace estimation.}\label{sec:appendix_geometric_assumptions}
The activation difference $\Delta_o^{(l)}(x_i)$ asks a more specific question than standard probing: not only whether an NLI label is decodable from a representation, but whether applying the same semantic operation induces a consistent geometric effect across examples. For each model, component, and layer, activation deltas $\Delta_o^{(l)} \in \mathbb{R}^{N_o \times d}$ are computed as the difference between corrupted and clean activations. The operation subspace $\mathcal{S}_o^{(l)}$ is estimated via truncated SVD applied to the mean-centred delta matrix, retaining the top $k{=}4$ right singular vectors as an orthonormal basis. This rank is fixed uniformly across all operations, layers, components, and models; sensitivity was verified qualitatively at $k \in \{2, 8\}$, yielding no substantive change in relative ordering. Up to $N{=}300$ paired examples are used per operation; all layers are extracted at stride 1, covering indices $0$ through $L{-}1$.

Two assumptions underlie this procedure. \textbf{Assumption 1: Paired isolation} (see App.~\ref{app:atomicity} and App.~\ref{sec:appendix_data_generation} for the formal statement and the generation protocol that enforces it). \textbf{Assumption 2: Low-rank structure}: variability in activation deltas is assumed to be approximately low-rank, with the transformation effect concentrated in a small number of directions relative to $d$. The true intrinsic dimensionality of each operation's representation is unknown and may vary across operations and model families. \textbf{Assumption 3: Linearity}: the SVD framework and linear intervention assume that transformation effects are approximately linear in representation space; nonlinear displacements are captured only in projection, potentially underestimating operation-specific structure.

\paragraph{Random-subspace baselines.}
To confirm that SVD-derived subspaces capture operation-specific structure rather than arbitrary projections, each subspace metric is compared against a random-baseline control. Random orthonormal subspaces of rank $k{=}4$ are generated by drawing a Gaussian matrix $G \in \mathbb{R}^{d \times k}$ with a layer-dependent seed, then applying QR decomposition: $G = QR$, $Q \in \mathbb{R}^{d \times k}$. Five independent random seeds are evaluated per layer, and the mean classification accuracy across seeds and layers is reported as the chance-level reference.

\paragraph{Classification.}
Linear classifiers are evaluated using 5-fold stratified cross-validation (\texttt{random\_seed}{=}42), with estimation and evaluation splits fully disjoint across folds. Three classifiers are compared at each layer: logistic regression ($C{=}0.1$, \texttt{liblinear} solver, balanced class weights, \texttt{max\_iter}{=}1000), linear SVM ($C{=}0.1$, balanced class weights), and RBF-kernel SVM ($C{=}1.0$). The best-layer accuracy reported for each model-component combination is the maximum mean accuracy across layers for the logistic regression classifier, consistent with the classification results in Table~\ref{tab:classification}.

\paragraph{Selectivity permutation test.}
Statistical significance of the selectivity ratio $\rho_o^{(l)}$ is assessed using a one-sided permutation test over the selectivity matrix. The observed statistic is the mean on-diagonal projection magnitude. Under each of $B{=}5{,}000$ permutations, the column indices of the selectivity matrix are randomly shuffled, and the mean of the resulting pseudo-diagonal is recorded as the null distribution. The empirical $p$-value is computed as $(|\{b : \text{null}_b \geq \text{observed}\}| + 1) / (B + 1)$.

\paragraph{Steering hyperparameters.}
Activation steering is applied at all layers using the residual stream component (\texttt{resid\_post}). The top singular vector of the SVD subspace (index 0) is used as the SVD steering direction; the mean-shift direction (CAA) is computed as the arithmetic mean of the per-example deltas for each operation. Steering magnitude is swept over $\alpha \in \{5, 10, 20\}$ for the main evaluation, and over $\alpha \in \{0.5, 1.0, 2.0\}$ for the sensitivity analysis reported in Appendix~\ref{sec:app_steering_per_direction}. 

\paragraph{CAA vs SVD comparison permutation test.}
The significance of the direction-type advantage (CAA over SVD) is assessed using a one-sided paired sign-flip permutation test. For each model and regime, the per-operation differences $d_o = \text{flip\_to\_target}_\text{CAA} - \text{flip\_to\_target}_\text{SVD}$ are computed at the best $\alpha$ per layer. Under each of $B{=}10{,}000$ permutations, each $d_o$ is independently multiplied by a random sign $s_o \in \{-1, +1\}$ drawn uniformly, and the mean of the permuted differences is recorded. The $p$-value is $(|\{b : \bar{d}_b \geq \bar{d}_\text{obs}\}| + 1) / (B + 1)$, where $\bar{d}_\text{obs}$ is the observed mean difference.

\paragraph{Random-baseline steering directions.}
Each structured steering vector (SVD or CAA) is matched against a random unit-norm direction of the same dimensionality, scaled to the same $\ell_2$ norm as the structured vector. Three independent random seeds are used per evaluation to estimate the null-distribution flip rate. The contamination metric for cross-operation interference is defined as the difference between the observed and random-baseline flip-to-other rates: $\text{contamination} = \text{flip\_to\_other} - \text{rand\_flip\_to\_other}$, where positive off-diagonal contamination indicates that the structured direction disrupts another operation's predictions more than a random direction of equal magnitude.

\subsection{Additional Analysis Methods}
\label{sec:app_additional_methods}

\subsubsection{Layer-wise emergence}
\label{sec:app_layerwise}
We track selectivity $\rho_o^{(l)}$ and classification accuracy as a function of layer depth $l$ to identify where transformation structure emerges. Full layer-wise profiles are shown in App.~\ref{app:selectivity_curves_full} and Fig.~\ref{fig:classification_curves}.

\subsection{Model Specifications}
\label{sec:model_params}\label{sec:appendix_model_cards}
Table~\ref{tab:model_properties} summarises the Transformer models used in this study. We select four open-weight decoder-only models spanning the 1.7-4B parameter range: Qwen3-1.7B~\citep{qwen3technicalreport}, Qwen2.5-3B-Instruct (Qwen2.5-3B)~\citep{qwen2.5}, Phi-3-mini-4k-instruct (Phi-3)~\citep{abdin2024phi3}, and Gemma3-4B-Instruct (Gemma3-4B)~\cite{gemmateam2025gemma3}. The selection is governed by three criteria:
\begin{itemize}[leftmargin=*,itemsep=2pt]
    \item \textbf{Architectural comparability.} All four models are decoder-only Transformers while differing in training data composition, tokenisation, and optimisation regimes. This allows us to test whether transformation-level structure is consistent across model families rather than specific to a single implementation.
    \item \textbf{Controlled scale.} The 1.5-4B parameter range provides a regime in which models are sufficiently expressive to exhibit non-trivial semantic behaviour while remaining comparable in depth and width. This reduces confounding effects from large differences in capacity and facilitates layer-wise analysis at matched relative depths.
    \item \textbf{Pretraining distribution.} Qwen models are trained on
    multilingual corpora, while Phi-3 and Gemma3-4B are predominantly
    English-focused. This variation allows us to test whether the
    emergence of transformation-level structure depends on linguistic
    diversity in the training data.

\end{itemize}

All models are used in their released inference configurations and are analysed without fine-tuning. Licences are as follows: Qwen models are released under the Qwen licence agreement, Gemma under the Gemma Terms of Use, and Phi under the MIT licence. All experiments are conducted in accordance with the intended research use of these models.

\begin{table}
\centering
\caption{Model properties across architectures. Params: number of parameters, Layers: number of layers, D\textsubscript{model}: size of word embeddings and hidden states, Heads: number of attention heads, Act.: Activation function, MLP Dim: dimensionality of the FF layers.}
\resizebox{\columnwidth}{!}{%
\begin{tabular}{|l|c|c|c|c|c|c|}
\hline
\textbf{Model} & \textbf{Params} & \textbf{Layers} & \textbf{D\textsubscript{model}} & \textbf{Heads} & \textbf{Act.} & \textbf{MLP Dim} \\
\hline
gemma-3-4b-it & 3.4B & 34 & 2560 & 8 & gelu\_pytorch\_tanh & 10240          \\
phi-3         & 3.6B & 32 & 3072 & 32 & SiLU & 8192          \\
Qwen2.5-3b-it & 3.0B & 36 & 2048 & 16 & SiLU & 8192          \\
Qwen3-1.7B & 1.5B & 28 & 2048 & 16 & SiLU & 2048          \\
\hline
\end{tabular}%
}
\label{tab:model_properties}
\end{table}

\subsection{Software and Implementation}\label{app:software}
Experiments were conducted on H100 GPU, using Python~3.12 with the following core dependencies: NumPy~1.26.4, pandas~2.2.2, PyTorch~2.10.0~\citep{paszke2019pytorch}, TransformerLens~2.18.0~\citep{nanda2022transformerlens}, Transformers~4.57.6, scikit-learn~1.7.0~\cite{scikit-learn}, SciPy~1.15.3.

\section{Additional Results}\label{app:full-results}

\subsection{Selectivity}\label{app:selectivity_curves_full}
\begin{figure}[t]
    \centering
    \includegraphics[width=\linewidth]{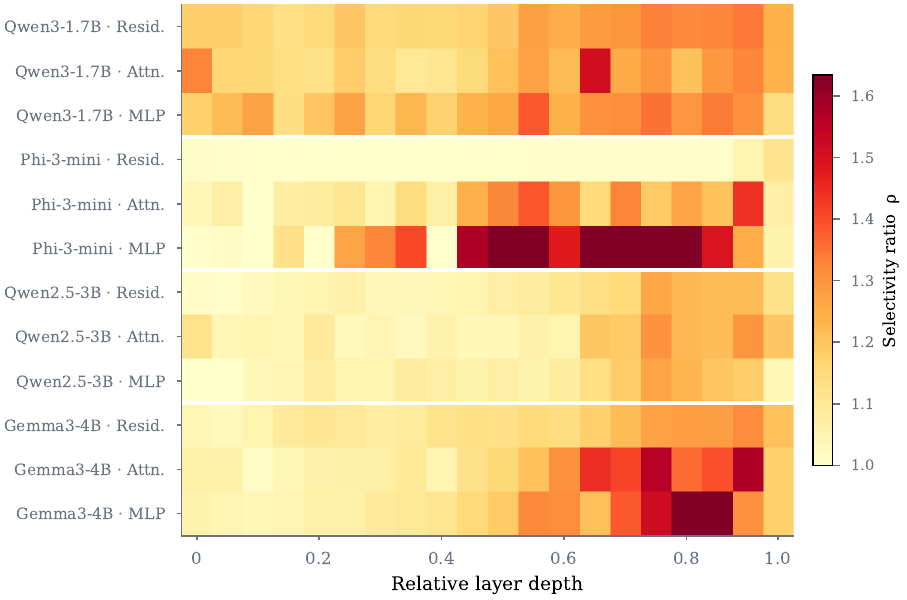}
    \caption{Selectivity ratio across relative layer depth for all model-component combinations. Rows correspond to model-component pairs and columns to relative layer depth. Warmer colours indicate higher selectivity. Selectivity increases toward the upper network across models, with strongest effects in MLP components.}
    \label{fig:fig5_selectivity_2d_heatmap}
\end{figure}

\begin{figure*}[h!] 
\centering 
\begin{subfigure}[t]{0.48\linewidth} 
\includegraphics[width=\linewidth]{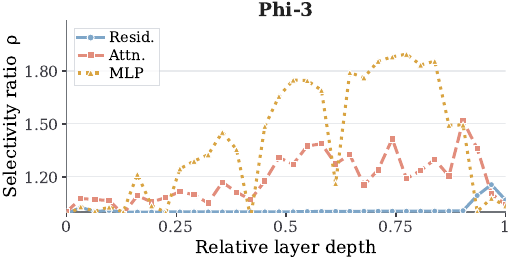} 
\end{subfigure} \hfill 
\begin{subfigure}[t]{0.48\linewidth}
\includegraphics[width=\linewidth]{figures/fig1_selectivity_curves_gemma_3_4b_it.pdf} 
\end{subfigure} 
\begin{subfigure}[t]{0.48\linewidth}
\includegraphics[width=\linewidth]{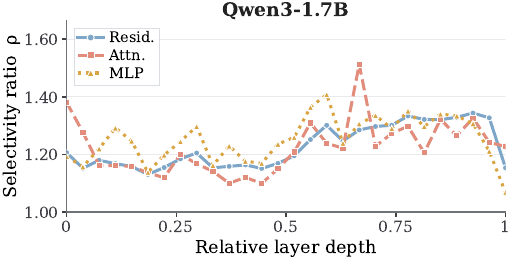} 
\end{subfigure} 
\begin{subfigure}[t]{0.48\linewidth}
\includegraphics[width=\linewidth]{figures/fig1_selectivity_curves_qwen2_5_3b_instruct.pdf} 
\end{subfigure} 
\caption{Layer-wise selectivity ratio $\rho$ across relative depth. Selectivity is low in early layers ($\rho \approx 1$) and increases toward the upper network, with strongest effects in MLP components.} 
\label{fig:fig1_selectivity_curves_full} 
\end{figure*}

\begin{figure*}[t]
    \centering
    \begin{subfigure}[t]{0.48\linewidth} 
    \includegraphics[width=\linewidth]{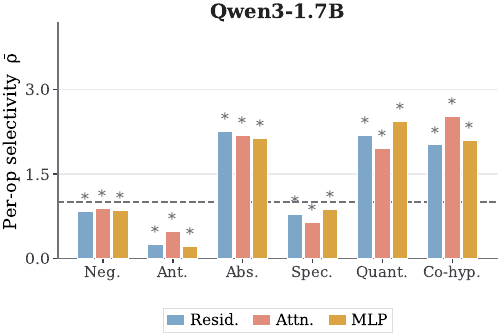} 
    \end{subfigure} \hfill 
    \begin{subfigure}[t]{0.48\linewidth} 
    \includegraphics[width=\linewidth]{ 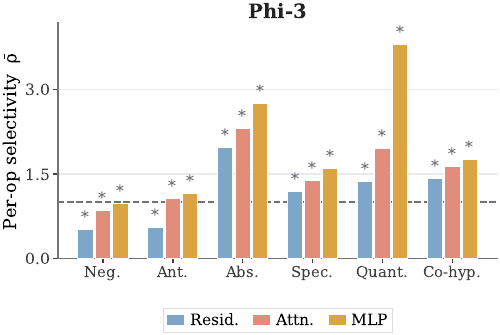} 
    \end{subfigure} \vspace{0.5em} 
    \begin{subfigure}[t]{0.48\linewidth} 
    \includegraphics[width=\linewidth]{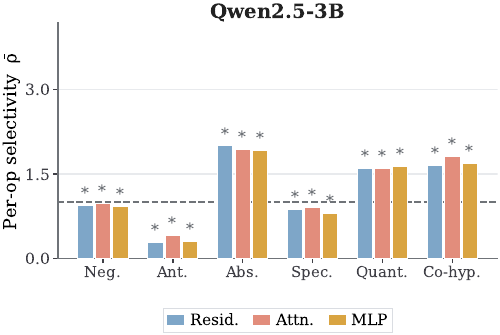}
    \end{subfigure} \hfill 
    \begin{subfigure}[t]{0.48\linewidth}
    \includegraphics[width=\linewidth]{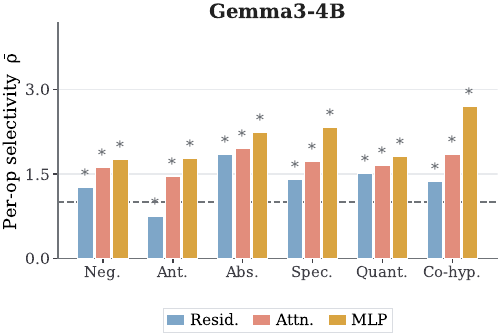} 
    \end{subfigure}
    \caption{Per-operation selectivity ratio $\rho$ at the peak-selectivity layer for all four models. {Abstraction} exhibits the highest selectivity across models and components, followed by {Quantification} and {Co-Hyponyms}, while {Negation} and {Antonymy} show $\rho < 1$ in several configurations. The main paper reports the Phi-3 and Gemma3-4B subset.}

    \label{app:fig_per_op_selectivity_full}
\end{figure*}

Fig.~\ref{fig:fig5_selectivity_2d_heatmap} shows layer-wise selectivity ratio $\rho$ across relative depth for all model-component combinations, while Fig.~\ref{fig:fig1_selectivity_curves_full} shows layer-wise selectivity curves for all four models and components. Tab.~\ref{tab:selectivity_full} gives the full numerical results. 
Fig.~\ref{app:fig_per_op_selectivity_full} shows per-operation selectivity ratios at the peak-selectivity layer for all four models. The heterogeneity described in \S\ref{sec:results} is consistent across models: {Abstraction} exhibits the highest per-operation selectivity across most model-component combinations, followed by {Quantification} and {Co-Hyponyms}, while {Negation} and {Antonymy} fall below $\rho = 1$ in several configurations.
\begin{table*}[t!]
\centering
\small
\setlength{\tabcolsep}{4pt}
\caption{%
  Full subspace selectivity results.  Layer: absolute layer index at which $\rho$ is maximised. Rel.: relative depth (layer\,/\,(total layers\,$-$\,1)). $\mu_{\mathrm{on}}$, $\mu_{\mathrm{off}}$: mean projection magnitudes of on-target and off-target activation deltas onto the operation subspace at the peak layer.  All permutation $p$-values computed with 1000 random orthonormal subspace draws from the Gaussian manifold.%
}
\label{tab:selectivity_full}
\begin{tabular}{llcccrrr}
\toprule
\textbf{Model} & \textbf{Component} &
  $\boldsymbol{\rho}$ & $\boldsymbol{d}$ &
  $\boldsymbol{p}$ &
  \textbf{Layer} & \textbf{Rel.} &
  $\boldsymbol{\mu_{\mathrm{on}}}$ / $\boldsymbol{\mu_{\mathrm{off}}}$ \\
\midrule
\multirow{3}{*}{Qwen3-1.7B}
  & \texttt{resid\_post}
    & 1.344 & 0.537 & 0.002 & 25 & 0.93
    & 166.8\,/\,124.2 \\
  & \texttt{attn\_out}
    & 1.513 & 0.658 & 0.001 & 18 & 0.67
    & 7.2\,/\,4.8 \\
  & \texttt{mlp\_out}
    & 1.408 & 0.607 & 0.002 & 16 & 0.59
    & 9.3\,/\,6.6 \\
\midrule
\multirow{3}{*}{Qwen2.5-3B}
  & \texttt{resid\_post}
    & 1.262 & 0.464 & 0.003 & 27 & 0.77
    & 7.3\,/\,5.8 \\
  & \texttt{attn\_out}
    & 1.304 & 0.569 & 0.001 & 27 & 0.77
    & 2.3\,/\,1.8 \\
  & \texttt{mlp\_out}
    & 1.267 & 0.451 & 0.004 & 27 & 0.77
    & 5.0\,/\,4.0 \\
\midrule
\multirow{3}{*}{Phi-3}
  & \texttt{resid\_post}
    & 1.155 & 0.353 & 0.002 & 30 & 0.97
    & 24.4\,/\,21.1 \\
  & \texttt{attn\_out}
    & 1.518 & 1.111 & 0.001 & 28 & 0.90
    & 2.4\,/\,1.6 \\
  & \texttt{mlp\_out}
    & 1.896 & 1.233 & 0.005 & 24 & 0.77
    & 1.2\,/\,0.7 \\
\midrule
\multirow{3}{*}{Gemma3-4B}
  & \texttt{resid\_post}
    & 1.346 & 1.095 & 0.002 & 31 & 0.94
    & 1879.5\,/\,1396.5 \\
  & \texttt{attn\_out}
    & 1.701 & 2.154 & 0.001 & 31 & 0.94
    & 482.1\,/\,283.4 \\
  & \texttt{mlp\_out}
    & \textbf{2.072} & \textbf{2.402} & 0.002 & 27 & 0.82
    & 234.1\,/\,113.0 \\
\bottomrule
\end{tabular}
\end{table*}


\subsection{Classification}\label{sec:appendix_classification}\label{app:classification_results}
\begin{figure*}[t]
\centering
\includegraphics[width=\linewidth]{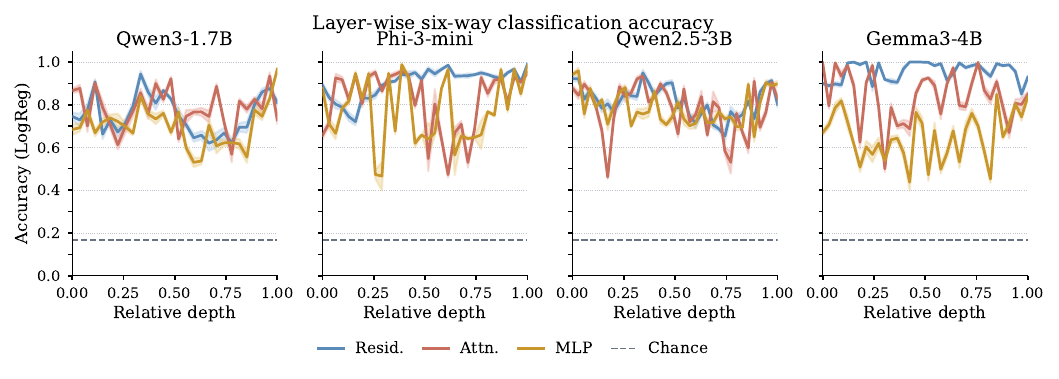}
\caption{Layer-wise six-way classification accuracy (logistic regression) across relative depth for all four models and three components. Shaded bands show $\pm$1 SD across five cross-validation folds. Dashed line marks chance ($1/6 \approx 0.167$). Accuracy rises sharply in early-to-mid layers and plateaus in upper layers, consistent with the selectivity profile in Fig.~\ref{fig:fig1_selectivity_curves}.}
\label{fig:classification_curves}
\end{figure*}

Tab.~\ref{tab:classification} reports best-layer linear classification accuracy for the six-way operation classification task across all model-component combinations. All configurations exceed the chance level of $1/6 \approx 16.7\%$ by a substantial margin, with most reaching above $93\%$. The best-layer for classification does not always coincide with the best-layer for selectivity, indicating that the two metrics assess different but complementary aspects of the representational geometry. Fig.~\ref{fig:classification_curves} shows the full layer-wise accuracy profiles.

\begin{table}[t]
\centering
\small
\caption{Best-layer linear classification accuracy for six-way operation classification (chance $= 1/6 = 0.167$). Mean and standard deviation are computed across five stratified cross-validation folds on held-out deltas projected into the per-operation subspaces.}
\label{tab:classification}
\resizebox{.5\textwidth}{!}{
\begin{tabular}{llccc}
\toprule
\textbf{Model} & \textbf{Component} & \textbf{Layer} & \textbf{Accuracy} & \textbf{SD} \\
\midrule
\multirow{3}{*}{Qwen-1.7B}
  & \texttt{resid\_post} & 9  & 0.944 & 0.015 \\
  & \texttt{attn\_out}   & 26 & 0.934 & 0.017 \\
  & \texttt{mlp\_out}    & 27 & 0.966 & 0.005 \\
\midrule
\multirow{3}{*}{Qwen2.5-3B}
  & \texttt{resid\_post} & 12 & 0.949 & 0.015 \\
  & \texttt{attn\_out}   & 12 & 0.934 & 0.012 \\
  & \texttt{mlp\_out}    & 1  & 0.959 & 0.015 \\
\midrule
\multirow{3}{*}{Phi-3}
  & \texttt{resid\_post} & 31 & 0.989 & 0.005 \\
  & \texttt{attn\_out}   & 12 & 0.956 & 0.007 \\
  & \texttt{mlp\_out}    & 12 & 0.987 & 0.005 \\
\midrule
\multirow{3}{*}{Gemma-4B}
  & \texttt{resid\_post} & 7  & 0.999 & 0.001 \\
  & \texttt{attn\_out}   & 25 & 0.998 & 0.003 \\
  & \texttt{mlp\_out}    & 28 & 0.848 & 0.014 \\
\bottomrule
\end{tabular}
}
\end{table}

\subsection{Clustering}\label{sec:appendix_tsen}
Silhouette scores are computed on t-SNE projections of activation deltas at the best-silhouette layer for each model-component combination (Tab.~\ref{tab:silhouette}). All twelve scores are negative, indicating that projected representations do not form well-separated clusters in two dimensions. This is consistent with the subspace geometry characterised in \S\ref{sec:results}: the full $d$-dimensional subspaces are linearly separable to a classifier (Tab.~\ref{tab:classification}), but two-dimensional projections do not yield sharp cluster boundaries, reflecting the graded overlap in the selectivity ratio profiles. Silhouette scores are treated as a convergent geometric diagnostic rather than primary evidence.

\begin{table}[t]
\centering
\small
\caption{Silhouette scores on t-SNE projections of activation deltas at the best-silhouette layer per model-component combination. All values are negative, confirming the absence of well-separated clusters in two-dimensional projection.}
\label{tab:silhouette}
\begin{tabular}{llcc}
\toprule
\textbf{Model} & \textbf{Component} & \textbf{Best layer} & \textbf{Silhouette} \\
\midrule
\multirow{3}{*}{Qwen3-1.7B}
  & \texttt{resid\_post} & 25 & $-0.176$ \\
  & \texttt{attn\_out}   &  3 & $-0.168$ \\
  & \texttt{mlp\_out}    & 25 & $-0.161$ \\
\midrule
\multirow{3}{*}{Phi-3}
  & \texttt{resid\_post} & 31 & $-0.180$ \\
  & \texttt{attn\_out}   & 29 & $-0.142$ \\
  & \texttt{mlp\_out}    & 30 & $-0.186$ \\
\midrule
\multirow{3}{*}{Qwen2.5-3B}
  & \texttt{resid\_post} & 31 & $-0.230$ \\
  & \texttt{attn\_out}   & 32 & $-0.038$ \\
  & \texttt{mlp\_out}    & 33 & $-0.205$ \\
\midrule
\multirow{3}{*}{Gemma3-4B}
  & \texttt{resid\_post} & 31 & $-0.127$ \\
  & \texttt{attn\_out}   & 31 & $-0.051$ \\
  & \texttt{mlp\_out}    & 31 & $-0.134$ \\
\bottomrule
\end{tabular}
\end{table}

\subsection{Logit Lens Trajectories} \label{app:logit_lens_full}
\begin{figure*}[h] 
    \centering 
    \begin{subfigure}[t]{0.49\linewidth}  
        \includegraphics[width=\linewidth]{ 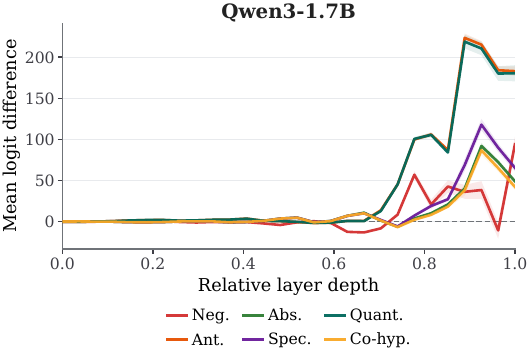} 
    \end{subfigure} \hfill 
    \begin{subfigure}[t]{0.49\linewidth} \includegraphics[width=\linewidth]{ figures/fig4_logit_lens_phi_3.pdf} 
    \end{subfigure} \vspace{0.5em} 
    \begin{subfigure}[t]{0.49\linewidth} \includegraphics[width=\linewidth]{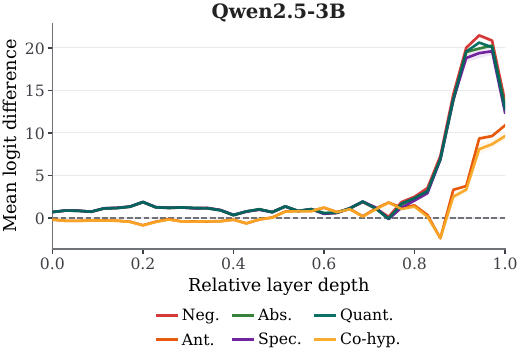} \end{subfigure} \hfill 
    \begin{subfigure}[t]{0.49\linewidth} \includegraphics[width=\linewidth]{figures/fig4_logit_lens_gemma_3_4b_it.pdf} 
    \end{subfigure} 
    \caption{Logit lens trajectories across relative depth for all four models and all six operations. Positive values indicate that the decoded residual stream favours the target label induced by the transformation. Qwen3-1.7B, Phi-3, and Qwen2.5-3B show near-zero trajectories through early and middle layers, followed by sharp amplification in the upper network. Gemma3-4B exhibits a three-phase pattern: gradual early rise, mid-network suppression, and strong late amplification with substantially larger magnitudes.} 
    \label{fig:logit_lens_full} 
\end{figure*}

Fig.~\ref{fig:logit_lens_full} shows layer-wise logit difference trajectories for all four models. Qwen3-1.7B and Qwen2.5-3B follow the same late-amplification pattern described in \S\ref{sec:results}: near-zero trajectories through early and middle layers, with sharp amplification concentrated in the final $10$-$15\%$ of layers.

\subsection{Direct Logit Attribution}\label{app:dla}
\begin{figure}[t]
    \centering
    \includegraphics[width=\linewidth]{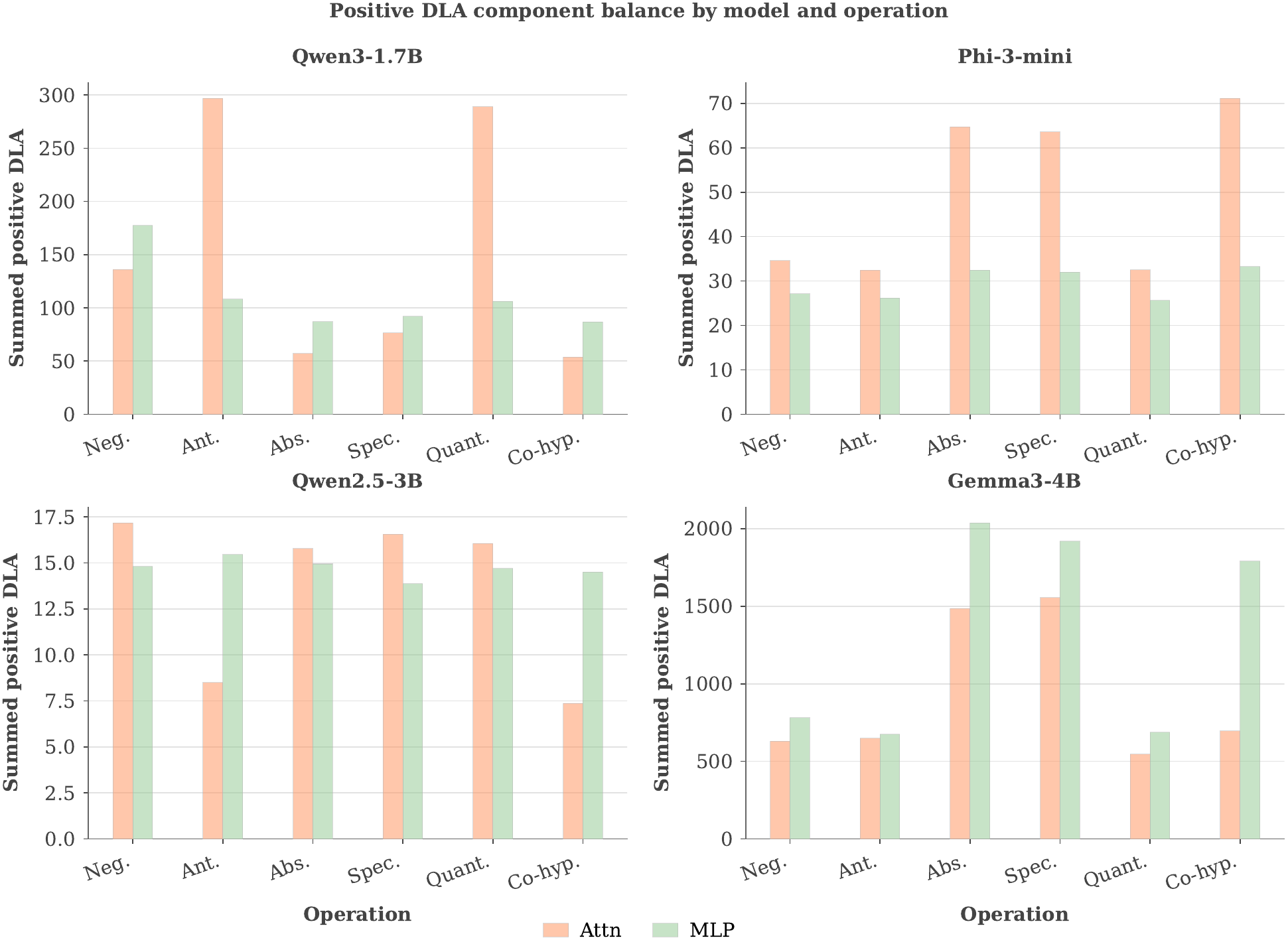}
    \caption{
Summed positive direct logit attribution from attention and MLP components for each operation and model. Component contributions vary across architectures and operations, with no single component type uniformly dominant.
}
    \label{fig:rq4_dla_component_balance}
\end{figure}

Direct logit attribution (DLA) decomposes the final logit difference into additive contributions from attention and MLP components at each layer (\S\ref{sec:method_localisation}). Fig.~\ref{fig:rq4_dla_component_balance} shows summed positive DLA per operation and model. The dominant contributing component type varies across model families: Phi-3 is predominantly attention-driven, Gemma3-4B is MLP-dominant for most operations, Qwen3-1.7B shows mixed operation-dependent contributions, and Qwen2.5-3B exhibits relatively balanced effects. No single component type is uniformly responsible across all models and operations.

\subsection{Per-Direction Steering Results}\label{sec:app_steering_per_direction}

\begin{figure*}[t]
    \centering
    \includegraphics[width=\linewidth]{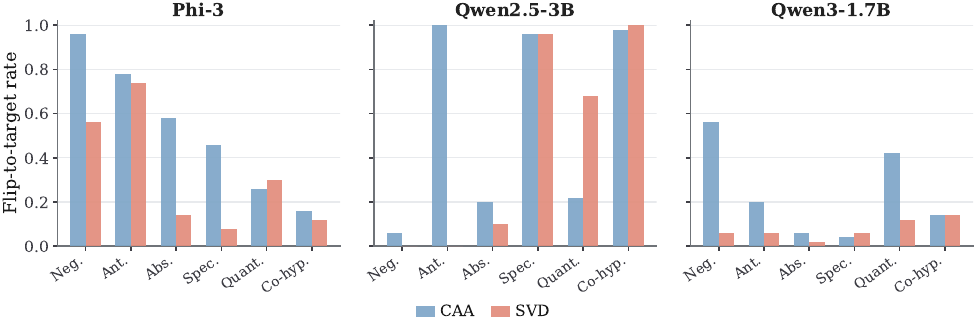}
    \includegraphics[width=\linewidth]{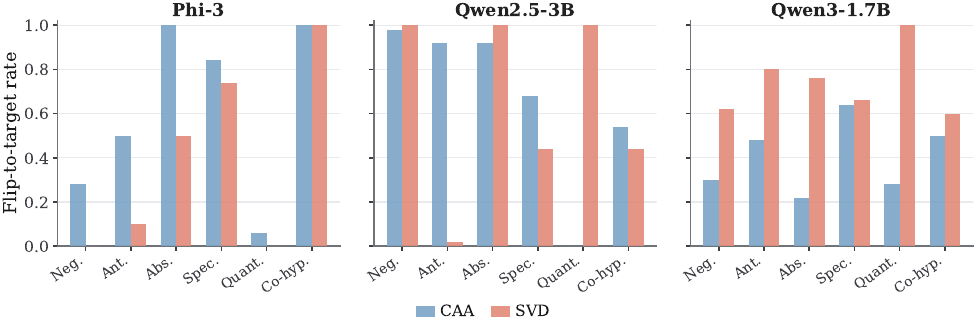}
    \caption{
    Per-operation flip-to-target rates for CAA mean-shift directions and SVD principal variance directions, shown separately for forward (induction; top) and inverse (removal; bottom) steering regimes. Each bar pair compares the two direction types for one operation in one model. Gemma3-4B is omitted because both direction types yield uniformly near-zero flip rates.
    }
    \label{fig:fig_caa_svd_comparison}
\end{figure*}

Fig.~\ref{fig:fig_caa_svd_comparison} shows the underlying per-operation flip-to-target rates. The full per-comparison breakdown is provided in Tab.~\ref{tab:caa_svd_full}.
A forward-inverse asymmetry is shown. In the forward regime, CAA directions achieve a non-tied win rate of $0.750$ ($12$ wins from $16$ non-tied comparisons across three steerable models), with SVD winning in $4$ cases: Quantification for Phi-3 and Qwen2.5-3B, Co-hyponymy for Qwen2.5-3B, and Specification for Qwen3-1.7B. Three of these four SVD advantages are concentrated in lower-alignment operations (Quantification: $\cos = 0.459$; Co-hyponymy: $\cos = 0.514$), consistent with the principal variance direction providing a more concentrated causal signal when mean-shift and variance axes diverge substantially; the exception is Qwen3-1.7B Specification ($\cos = 0.810$, narrow margin: SVD $0.06$, CAA $0.04$). In the inverse regime, the non-tied win rate is $0.471$ ($8$ CAA wins, $9$ SVD wins from $17$ non-tied comparisons), with a slight SVD advantage overall. The most striking pattern is a complete forward-inverse reversal for Qwen3-1.7B: CAA leads in the forward regime (win rate $0.667$) but SVD wins all six operations in the inverse regime (win rate $0.000$, $\bar{\Delta} = -0.337$), suggesting that the directions that best induce transformation effects are not the same as those that best remove them. SVD wins in the inverse regime are concentrated in Qwen3-1.7B (six wins) and Qwen2.5-3B (three wins, including Quantification: SVD $1.00$, CAA $0.00$), while Phi-3 inverse steering is CAA-dominant across all operations.
\begin{table}[t]
\centering
\small
\caption{Full per-operation, per-model CAA vs SVD steering comparison. SVD and CAA columns report flip-to-target rates at the best $\alpha$ across all layers. Tie\_zero: both directions produce zero. Tie\_nonzero: both directions produce the same nonzero rate. Gemma3-4B rows are all Tie\_zero and are omitted for space.}
\label{tab:caa_svd_full}
\resizebox{.5\textwidth}{!}{
\begin{tabular}{llccl}
\toprule
\textbf{Model} & \textbf{Operation} & \textbf{SVD} & \textbf{CAA} & \textbf{Winner} \\
\midrule
\multicolumn{5}{l}{\textit{Forward regime}} \\
\midrule
\multirow{6}{*}{Phi-3}
  & Abstraction    & 0.14 & 0.58 & CAA \\
  & Antonymy       & 0.74 & 0.78 & CAA \\
  & Co-hyponymy    & 0.12 & 0.16 & CAA \\
  & Negation       & 0.56 & 0.96 & CAA \\
  & Quantification & 0.30 & 0.26 & SVD \\
  & Specification  & 0.08 & 0.46 & CAA \\
\midrule
\multirow{6}{*}{Qwen2.5-3B}
  & Abstraction    & 0.10 & 0.20 & CAA \\
  & Antonymy       & 0.00 & 1.00 & CAA \\
  & Co-hyponymy    & 1.00 & 0.98 & SVD \\
  & Negation       & 0.00 & 0.06 & CAA \\
  & Quantification & 0.68 & 0.22 & SVD \\
  & Specification  & 0.96 & 0.96 & Tie (nonzero) \\
\midrule
\multirow{6}{*}{Qwen3-1.7B}
  & Abstraction    & 0.02 & 0.06 & CAA \\
  & Antonymy       & 0.06 & 0.20 & CAA \\
  & Co-hyponymy    & 0.14 & 0.14 & Tie (nonzero) \\
  & Negation       & 0.06 & 0.56 & CAA \\
  & Quantification & 0.12 & 0.42 & CAA \\
  & Specification  & 0.06 & 0.04 & SVD \\
\midrule
\multicolumn{5}{l}{\textit{Inverse regime}} \\
\midrule
\multirow{6}{*}{Phi-3}
  & Abstraction    & 0.50 & 1.00 & CAA \\
  & Antonymy       & 0.10 & 0.50 & CAA \\
  & Co-hyponymy    & 1.00 & 1.00 & Tie (nonzero) \\
  & Negation       & 0.00 & 0.28 & CAA \\
  & Quantification & 0.00 & 0.06 & CAA \\
  & Specification  & 0.74 & 0.84 & CAA \\
\midrule
\multirow{6}{*}{Qwen2.5-3B}
  & Abstraction    & 1.00 & 0.92 & SVD \\
  & Antonymy       & 0.02 & 0.92 & CAA \\
  & Co-hyponymy    & 0.44 & 0.54 & CAA \\
  & Negation       & 1.00 & 0.98 & SVD \\
  & Quantification & 1.00 & 0.00 & SVD \\
  & Specification  & 0.44 & 0.68 & CAA \\
\midrule
\multirow{6}{*}{Qwen3-1.7B}
  & Abstraction    & 0.76 & 0.22 & SVD \\
  & Antonymy       & 0.80 & 0.48 & SVD \\
  & Co-hyponymy    & 0.60 & 0.50 & SVD \\
  & Negation       & 0.62 & 0.30 & SVD \\
  & Quantification & 1.00 & 0.28 & SVD \\
  & Specification  & 0.66 & 0.64 & SVD \\
\midrule
\multicolumn{4}{l}{\textbf{Forward non-tied win rate}} &
  CAA 0.750 / SVD 0.250 \\
\multicolumn{4}{l}{\textbf{Inverse non-tied win rate}} &
  CAA 0.471 / SVD 0.529 \\
\multicolumn{4}{l}{\textbf{Overall non-tied win rate (3 models)}} &
  CAA 0.606 / SVD 0.394 \\
\bottomrule
\end{tabular}
}
\end{table}

\subsection{CAA vs SVD Direction Alignment}\label{sec:app_caa_svd_sim}
\begin{figure}[t]
\centering
\includegraphics[width=\linewidth]{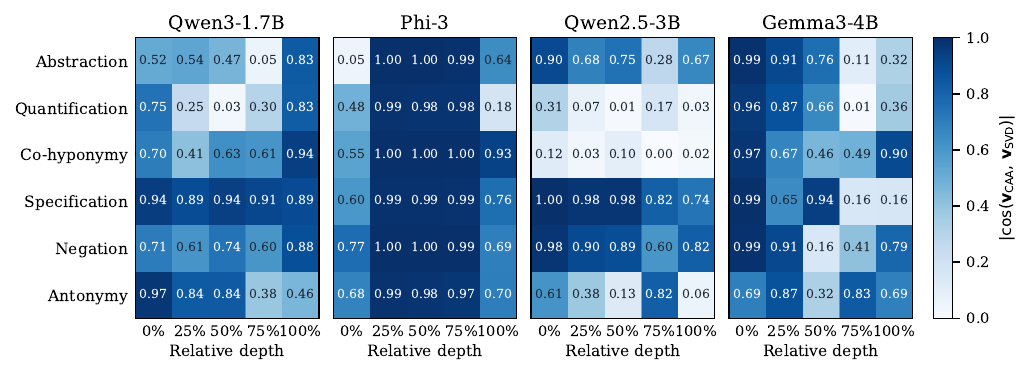}
\caption{Layer-wise cosine similarity between CAA and SVD steering directions at relative depth percentiles 0/25/50/75/100\%. Rows = operations (sorted by mean selectivity), columns = depth percentile, colour intensity = $|\cos(\mathbf{v}_\mathrm{CAA}, \mathbf{v}_\mathrm{SVD})|$.}
\label{fig:app_caa_svd_sim}
\end{figure}

Fig.~\ref{fig:app_caa_svd_sim} shows the cosine similarity between CAA (mean-shift) and SVD (top singular vector) steering directions at five relative depth percentiles (0, 25, 50, 75, 100\%) for each operation and model. Alignment varies substantially across models and operations, and evolves with depth. Phi-3 shows consistently high alignment across operations and layers, while Qwen2.5-3B shows strong operation-specific divergence (e.g.\ Co-hyponymy and Quantification remain near zero throughout). Early layers tend to show lower alignment, with convergence increasing toward the upper network for most configurations.

\subsection{Cross-Operation Interference: Per-Model and Inverse-Regime Results}\label{sec:app_cross_op}
\begin{figure*}[t]
    \centering
    \includegraphics[width=\linewidth]{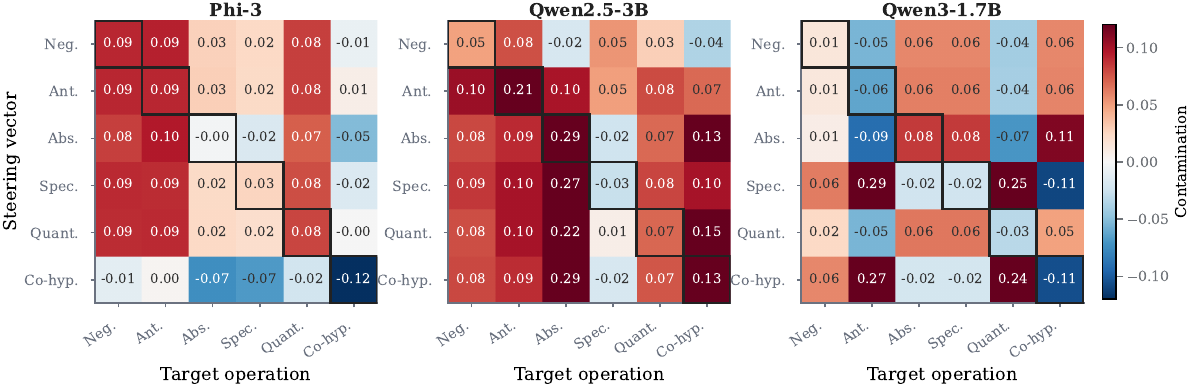}
    \caption{Per-model cross-operation steering contamination in the forward regime at $\alpha{=}20$, averaged over layers. Each panel corresponds to one model (three steerable models and Gemma3-4B). Rows index the source operation from which the SVD steering direction is derived; columns index the target operation whose held-out inputs are evaluated. Cell values give the excess flip-to-target rate relative to a matched random-direction baseline. The aggregate heatmap in the main paper (Fig.~\ref{fig:fig3_cross_op_heatmap}) reports the mean over all four models.}
    \label{fig:app_cross_op_permodel}
\end{figure*}

\begin{figure*}[t]
    \centering
    \includegraphics[width=\linewidth]{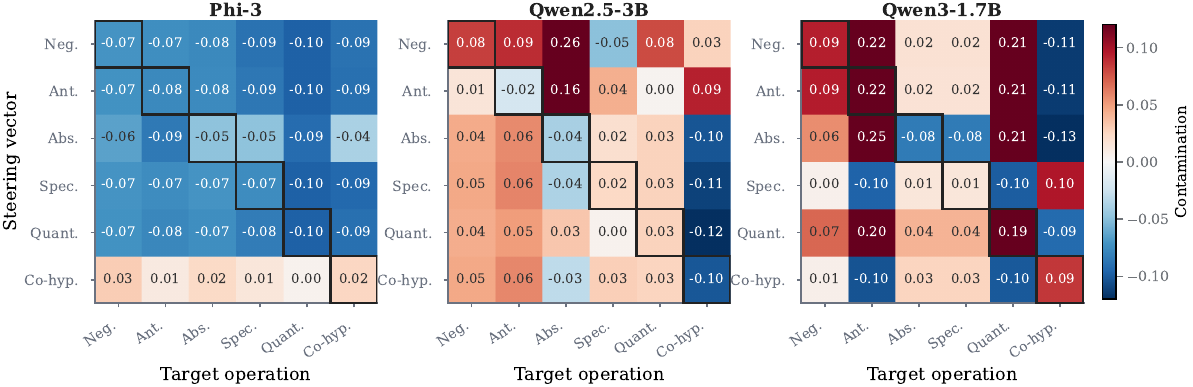}
    \caption{Per-model cross-operation steering contamination in the inverse regime at $\alpha{=}20$, averaged over layers. Patterns are more variable across models than in the forward regime, and no consistent source~$\to$~target interference structure is observed at the aggregate level.}
    \label{fig:app_cross_op_inverse}
\end{figure*}

Fig.~\ref{fig:app_cross_op_permodel} shows per-model cross-operation contamination heatmaps in the forward regime at $\alpha{=}20$, averaged over layers. The five high-interference pairs identified in the main text (contamination $> 0.07$) are model-dependent in their distribution. Qwen3-1.7B shows the most concentrated interference ($6$ of $30$ off-diagonal pairs above $0.07$), with the strongest effects involving Specification as source. Phi-3 shows broader interference ($12$ of $30$ pairs above $0.07$), concentrated among label-related pairs (Negation, Antonymy, Abstraction targeting Antonymy). Qwen2.5-3B exhibits the most diffuse pattern ($20$ of $30$ pairs above $0.07$), consistent with its weaker and more uniform per-operation selectivity profile in RQ1. Gemma3-4B shows near-zero contamination throughout the forward regime, reflecting the absence of reliable additive steering signal for this model. The inverse regime (Fig.~\ref{fig:app_cross_op_inverse}) exhibits more variable patterns across models, with no consistent source~$\to$~target structure at the aggregate level. We do not make strong structural claims about inverse-regime interference given this variability; the forward-regime pattern is the primary empirical basis for the RQ3 dissociation argument.

\subsection{Label-Controlled Operation Analysis}
\label{app:label_controlled_operations}

A possible confound in our setting is that operation identity is partially aligned with NLI label. Negation, Antonymy, and Quantification should produce contradiction examples, while Abstraction, Specification, and Co-hyponymy neutral examples. A direction that separates contradiction from neutral could therefore appear operation-selective even if it encodes only label-level structure. We address this directly with label-controlled analyses that test whether operation-specific structure remains when label identity is held fixed or explicitly removed.

\paragraph{Method.}
We use the same activation-difference and subspace notation as in \S\ref{sec:method_subspaces}. For operation $o$ at layer $l$, let $\mathbf{D}_o^{(l)} \in \mathbb{R}^{N \times d}$ denote the matrix of activation differences, and let $\mathbf{B}_o^{(l)} \in \mathbb{R}^{k \times d}$ be the row-stacked rank-$k$ basis obtained by mean-centred SVD. Projection onto the operation subspace is defined as $eratorname{proj}_{\mathcal{S}_o^{(l)}}(\boldsymbol{\delta}) = {\mathbf{B}_o^{(l)}}^\top \mathbf{B}_o^{(l)} \boldsymbol{\delta}$. Operations are grouped by label into contradiction and neutral sets. For each group, balanced samples are constructed by drawing equal numbers of activation differences from each operation, with replacement where necessary:
\begin{equation}
eratorname{proj}_{\mathcal{S}_o^{(l)}}(\boldsymbol{\delta})
= {\mathbf{B}_o^{(l)}}^\top \mathbf{B}_o^{(l)} \boldsymbol{\delta}.
\end{equation}
Operations are grouped by label into contradiction and neutral sets. For each group, balanced samples are constructed by drawing equal numbers of activation differences from each operation, with replacement where necessary.

We apply four controls. (i)~\emph{Within-label operation prediction}: operation identity is predicted while holding the NLI label fixed. (ii)~\emph{Label residualisation}: label-aligned components are removed by projecting activation differences orthogonally to a label subspace estimated via SVD. (iii)~\emph{Probe-depth comparison}: layer-wise peaks for label and operation prediction are compared. (iv)~\emph{Label-aware contamination}: cross-operation interference is compared between same-label and cross-label pairs.

\paragraph{Within-label operation prediction.}
Operation identity is recoverable from activation differences even when all examples share the same NLI label. Averaged across all model-component configurations, contradiction-only accuracy is $0.539$ and neutral-only accuracy is $0.655$, both well above the within-label chance rate of $1/3$ (Tab.~\ref{tab:label_control_within_label}). A classifier restricted to label-level information cannot exceed chance within a fixed-label group; above-chance performance therefore indicates the presence of operation-specific structure.

\begin{table}[t]
\centering
\small
\caption{Within-label operation prediction accuracy. Each classifier distinguishes operations while holding the NLI label fixed. Chance is $1/3$ within each label group. Results are averaged over all twelve model-component configurations.}
\label{tab:label_control_within_label}
\begin{tabular}{lccc}
\toprule
\textbf{Label group} & \textbf{Chance} & \textbf{Mean acc.} & \textbf{Range} \\
\midrule
Contradiction & 0.333 & 0.539 & 0.417-0.687 \\
Neutral       & 0.333 & 0.655 & 0.523-0.780 \\
\bottomrule
\end{tabular}
\end{table}

\paragraph{Within-label selectivity.}
Within-label selectivity ratios exceed $\rho = 1$ across all operations, models, and components (mean $\rho = 1.154$, minimum $\rho = 1.003$). This indicates that operation-specific structure persists within fixed-label groups, although with reduced separation consistent with the overlapping geometry observed in RQ1.

\paragraph{Label residualisation.}
Operation and label subspaces are strongly aligned (mean overlap $0.868$), and label-subspace projections account for $87.8\%$ of activation-difference variance. Despite this alignment, operation-specific structure remains after removing the label component. Mean selectivity increases from $\rho = 1.185$ to $\rho = 2.841$, and all residualised values remain above $\rho = 1$ (Tab.~\ref{tab:label_residualisation_summary}). This indicates that removing the shared label-aligned component increases the concentration of variance within operation-specific directions.

\begin{table}[t]
\centering
\small
\caption{Effect of label-subspace residualisation on selectivity. Values are averaged across all operations, models, and components. Resid./raw reports the ratio $\rho_{\mathrm{resid}} / \rho_{\mathrm{raw}}$.}
\label{tab:label_residualisation_summary}
\resizebox{.5\textwidth}{!}{
\begin{tabular}{ccccc}
\toprule
\textbf{Raw $\rho$} & \textbf{Residualised $\rho$} &
\textbf{Resid./raw} & \textbf{Overlap} & \textbf{Label var.} \\
\midrule
1.185 & 2.841 & 241.0\% & 0.868 & 87.8\% \\
\bottomrule
\end{tabular}}
\end{table}

\paragraph{Probe-depth comparison.}
Peak depths for label prediction and operation prediction differ in some model-component configurations but coincide in others. For example, Phi-3 \texttt{attn} peaks at relative depth $0.968$ for label prediction and $0.387$ for operation prediction, while Qwen2.5-3B \texttt{resid} peaks at $0.971$ and $0.400$, respectively. Other configurations, including Gemma3-4B, show no separation. These differences are configuration-dependent and do not provide consistent evidence for depth-wise separation.

\paragraph{Label-aware contamination.}
If cross-operation interference were driven by label identity, same-label operation pairs would exhibit higher contamination than cross-label pairs. This pattern is not observed. Across all models, contamination levels are similar for same-label and cross-label pairs, with no statistically significant differences ($p \geq 0.721$) and small effect sizes (Tab.~\ref{tab:label_aware_contamination}). Cross-operation interference is therefore not primarily explained by shared label identity.

\begin{table}[t]
\centering
\small
\caption{Label-aware cross-operation contamination. Same-label pairs share the NLI label; cross-label pairs do not. No model shows a statistically reliable same-label amplification.}
\label{tab:label_aware_contamination}
\resizebox{\linewidth}{!}{
\begin{tabular}{lcccc}
\toprule
\textbf{Model} & \textbf{Same-label} & \textbf{Cross-label} &
  $\boldsymbol{d}$ & $\boldsymbol{p}$ \\
\midrule
Qwen3-1.7B &  0.036 &  0.054 & -0.150 & 0.721 \\
Phi-3      &  0.007 &  0.001 &  0.087 & 0.826 \\
Qwen2.5-3B &  0.100 &  0.100 & -0.004 & 0.991 \\
Gemma3-4B  & -0.004 & -0.005 &  0.023 & 0.950 \\
\bottomrule
\end{tabular}
}
\end{table}

\paragraph{Interpretation.}
The four controls converge on the same conclusion. Operation-specific structure is not reducible to NLI label structure alone. Operation identity remains predictable within fixed-label groups, selectivity increases after removal of label-aligned components, and interference patterns do not track label identity. Although operation and label subspaces are strongly aligned, reflecting the design of the dataset, this alignment does not eliminate transformation-specific structure.

\end{document}